\documentclass[pdflatex,sn-mathphys-num]{sn-jnl}

\usepackage{graphicx}%
\usepackage{multirow}%
\usepackage{amsmath,amssymb,amsfonts}%
\usepackage{amsthm}%
\usepackage{mathrsfs}%
\usepackage[title]{appendix}%
\usepackage{xcolor}%
\usepackage{textcomp}%
\usepackage{manyfoot}%
\usepackage{booktabs}%
\usepackage{algorithm}%
\usepackage{algorithmicx}%
\usepackage{algpseudocode}%
\usepackage{listings}%
\usepackage{hyperref}
\usepackage{url} 
\usepackage{subcaption}

\theoremstyle{thmstyleone}%
\theoremstyle{thmstyletwo}%

\theoremstyle{thmstylethree}%

\usepackage[numbers,sort&compress]{natbib}

\begin{document}

\title[Title]{Where Do Tokens Go? Understanding Pruning Behaviors in STEP at High Resolutions}

\author[1]{\fnm{Michal} \sur{Szczepanski}}\email{michal.szczepanski@cea.fr}\equalcont{These authors contributed equally to this work.}

\author*[1]{\fnm{Martyna} \sur{Poreba}}\email{martyna.poreba@cea.fr}
\equalcont{These authors contributed equally to this work.}

\author[2]{\fnm{Karim} \sur{Haroun}}\email{karim.haroun@etu.univ-cotedazur.fr}

\affil*[1]{\orgdiv{Université Paris-Saclay}, \orgname{CEA, List}, \orgaddress{
\postcode{F-91120}, \city{Palaiseau}, \country{France}}}

\affil[2]{\orgdiv{I3S}, \orgname{Université Côte d’Azur, CNRS}, 
\city{Sophia Antipolis}, \postcode{06900}, \country{France}}

\abstract{Vision Transformers (ViTs) achieve state-of-the-art performance in semantic segmentation but are hindered by high computational and memory costs. To address this, we propose STEP (SuperToken and Early-Pruning), a hybrid token-reduction framework that combines dynamic patch merging and token pruning to enhance efficiency without significantly compromising accuracy. At the core of STEP is dCTS, a lightweight CNN-based policy network that enables flexible merging into superpatches. Encoder blocks integrate also early-exits to remove high-confident supertokens, lowering computational load. We evaluate our method on high-resolution semantic segmentation benchmarks, including images up to $1024\times1024$, and show that when dCTS is applied alone, the token count can be reduced by a factor of 2.5 compared to the standard $16 \times 16$ pixel patching scheme. This yields a 2.6× reduction in computational cost and a 3.4× increase in throughput when using ViT-Large as the backbone. Applying the full STEP framework further improves efficiency, reaching up to a 4× reduction in computational complexity and a 1.7× gain in inference speed, with a maximum accuracy drop of no more than 2.0\%. With the proposed STEP configurations, up to 40\% of tokens can be confidently predicted and halted before reaching the final encoder layer.}


\maketitle

\section{Introduction}\label{Intro}

Vision Transformers (ViTs) have demonstrated strong performance in semantic segmentation tasks, primarily thanks to their capacity to capture long-range dependencies. Numerous strategies have been proposed to harness the full potential of ViTs in this context. One line of work focuses on designing transformer architectures specifically tailored for semantic segmentation \cite{wang2021pyramid, zheng2021rethinkingSETR}. For example, SETR \cite{zheng2021rethinkingSETR} views segmentation as a sequence-to-sequence prediction task, while the Pyramid Vision Transformer (PVT) \cite{wang2021pyramid} introduces a hierarchical structure to better capture spatial information. Another prevalent approach involves enhancing the transformer-based backbone \cite{Liu_2021_swin} or modifying the task-specific decoder \cite{strudel2021segmenter, zhang2021k, xie2021segformer}. SegFormer \cite{xie2021segformer} enhances segmentation performance by integrating pyramid features without relying on positional encodings. Segmenter \cite{strudel2021segmenter} introduces learnable class tokens that interact with the encoder output to generate masks in a data-dependent manner. SegViT \cite{zhang2022segvit} pushes the boundaries of self-attention through its attention-to-mask (ATM) module that directly predicts segmentation masks from attention maps. More recently, several works have proposed lightweight or alternative segmentation schemes that leverage the strength of pre-trained ViT backbones. EoMT \cite{kerssies2025eomt} introduces an encoder-only mask transformer that reuses a frozen ViT backbone and a lightweight mask head, demonstrating that ViTs inherently encode sufficient spatial information for segmentation without complex decoders. CCASeg \cite{Yoo_2025_WACV} proposes a convolutional cross-attention decoder that efficiently aggregates multi-scale context with reduced computational overhead. U-MixFormer \cite{Yeom_2025_WACV} presents a U-Net–like transformer architecture with mix-attention blocks, achieving competitive performance through efficient feature fusion. S4Former \cite{Hu_2024_CVPR} designs a semi-supervised ViT framework with patch-adaptive self-attention, achieving strong results with only partial label supervision.

Despite their strong performance, ViTs still pose significant computational challenges. A primary concern is the quadratic complexity of the self-attention mechanism, which scales poorly with image resolution. As input image size increases, both computational cost and memory consumption grow substantially, hindering the practical deployment of ViTs. Although various efforts have been made to improve their efficiency, achieving a balance between computational complexity, latency, and performance remains difficult, including quantization \cite{lin2022fqvit, Yuan22PTQ4ViT, li2023vit, huang2024quantization, 10.1007/978-981-96-2064-7_20} knowledge distillation \cite{tiny_vit, 10678046} and pruning. Key studies have demonstrated that these model compression approaches can significantly reduce both model size and computational cost, thereby enhancing the practicality of ViTs in large-scale applications. In this context, we propose SuperToken and Early-Pruning (STEP), a novel token reduction mechanism designed to enhance the efficiency of ViT for semantic segmentation. In contrast to conventional grid-based patch processing, this approach produces superpatches of varying sizes thanks to the proposed dCTS module, allowing the number of tokens to adapt to the complexity of the image content. Furthermore, STEP integrates an early-pruning strategy, in which certain tokens are masked and halted early in the network pipeline, thereby reducing the computational load in subsequent layers. This paper is an extended version of our conference
paper \cite{proust2025step}. We make several new contributions:
\begin{itemize}
  \item We conduct extensive experiments on an NVIDIA A100 GPU to evaluate the STEP mechanism integrated into state-of-the-art Transformer backbones (ViT-Large and ViT-Base), using SegViT as the decoder and widely recognized semantic segmentation benchmarks.
  \item We also demonstrate the potential of our framework on high-resolution images (up to $1024\times1024$) to assess its scalability for semantic segmentation. The model maintains competitive accuracy while significantly reducing computational complexity and inference time. Notably, to the best of our knowledge, this is the first attempt to evaluate a token pruning strategy in the context of high-resolution semantic segmentation.
  \item We provide more in-depth analyses, ablation studies, and visualizations.  
\end{itemize}

\section{Vision Transformer Pruning: Prior Work}\label{SOTA}

Vision Transformers traditionally partition an image into a uniform grid, treating each patch as an individual token. However, this fixed strategy overlooks the varying importance of different image regions depending on the task. For instance, recognizing fine details may require a high token density, whereas homogeneous areas can be represented with fewer tokens. This raises a key question: is it necessary to process the same number of tokens for each input image? Given the substantial computational cost of ViTs, reducing the number of tokens emerges as a natural and effective way to improve efficiency. When examining existing approaches, pruning techniques can be broadly categorized based on the level at which they operate. Some methods act at the patch-level to reduce redundancy before the input reaches the ViT backbone. Others focus on token-token pruning, eliminating tokens based on similarity or learned importance throughout the transformer layers. Effectively addressing these challenges requires advanced strategies that consider task-specific requirements, reliable token importance metrics, and retraining schemes to compensate for information loss. Importantly, excessive pruning may lead to the removal of critical content, degrading overall model performance. Striking a balance between computational efficiency and accuracy preservation remains a central challenge in token pruning for ViT.

\subsection{Patch-level pruning}
Patch-level pruning includes the aggregation of neighboring patches into larger, semantically consistent units. Some existing methods rely on learned mechanisms that dynamically predict which patches should be merged, typically using lightweight neural modules. For example, CTS \cite{lu2023cts} retains the naively sliced square image patches and merges locally the most similar ones. For this purpose, it employs a class-agnostic policy network to predict whether a group of 2$\times$2 neighboring patches belongs to the same class. If so, the patches are merged and represented by a shared token, thereby reducing the overall token count. An alternative idea is to use adaptive resolution or mixed-scale tokenization \cite{havtorn_msvit_2023, CFViT, ronen_vision_2023}. These approaches dynamically select token sizes or resolutions based on the input image content, while still relying on square-shaped patches. In MSViT \cite{havtorn_msvit_2023} a lightweight, four-layer MLP serves as a gating mechanism, making binary decisions on whether a region should be tokenized coarsely (with 32×32 pixel patches) or finely (with 16×16 pixel patches). CF-ViT \cite{CFViT} proposes a coarse-to-fine inference strategy. The model first performs inference on coarse-grained patches. If the confidence is low, only the informative regions identified via global class attention are re-processed at finer granularity. The Quadtree algorithm, integrated into the Quadformer model \cite{ronen_vision_2023}, is combined with a saliency scorer to adaptively partition the image into patches of varying sizes. Regions with higher saliency are represented at higher resolution, while less salient areas are processed at lower resolution. A different strategy for reducing the number of patches is patch pruning, which aims to retain only the most informative patches for the target task. This selective retention can be guided by learned importance measures, enabling the model to focus its computational resources on semantically relevant regions while discarding redundant or background information, as demonstrated by PaPr \cite{mahmud2024paprtrainingfreeonesteppatch}.

\subsection{Token-level pruning}
Token-level pruning typically operates at intermediate layers by removing or merging tokens based on their estimated importance. This usually takes place after one or more Transformer blocks, once sufficient contextual information has been aggregated to make an informed decision about which tokens are less informative or redundant for the downstream task. In contrast to patch-level, token pruning leverages the evolving semantic representations of tokens as they propagate through the network. A key component is the scoring mechanism used to evaluate the importance of each token. These techniques can broadly be categorized into learned and heuristic approaches. Learned token pruning methods \cite{rao2021dynamicvit, fayyaz2022ats, 10.1145/3534678.3539260, kong2022spvit, liang2022evit, 9879366, CP-ViT} incorporate trainable modules into the ViT architecture to assess token informativeness. In contrast, heuristic token pruning can be applied to the off-the-shelf ViTs, without further finetuning \cite{10030157, bolya2022tome, Tang23DToP, wu2023ppt}. Regardless of the technique used, the derived score determines which tokens are retained and which can be safely discarded or merged.

\subsubsection{Token discarding}
Token discarding refers to selectively removing tokens based on predefined importance scores or confidence measures.These methods can typically be divided into hard and soft pruning. In hard pruning \cite{fayyaz2022ats, 9879366, AS-ViT, CP-ViT, MARCHETTI2025127449, rao2021dynamicvit, Zero-Tprune} less important tokens are completely removed based on a predefined importance score. In contrast, soft pruning does not eliminate tokens entirely. Instead, it either aggregates less informative tokens into consolidated representations package token \cite{kong2022spvit, liang2022evit, evo-vit}, or halts their further processing once they reach a sufficient confidence level \cite{Tang23DToP, courdier2023paumerpatchpausingtransformer, LiuWACV, 10483924, Yin_2022_CVPR}. 

DToP \cite{Tang23DToP} and DoViT \cite{LiuWACV} both adopt the use of dynamic early-exit mechanisms that adaptively prune tokens based on confidence scores computed at intermediate layers, with DoViT adding a reconstruction module for spatial consistency.  A-ViT \cite{Yin_2022_CVPR} proposes to halt tokens using a cumulative sigmoid-based score derived from token embeddings. Among the methods that focus on generating and consolidating representative tokens, SP-ViT \cite{kong2022spvit} stands out by introducing an attention-based multi-head token selector. This module is inserted at multiple points in the network to rank tokens by importance, consolidate similar ones, and prune the least informative. Similarly, EViT \cite{liang2022evit} focuses on the progressive selection of informative tokens during training. It masks and fuses regions that represent the inattentive tokens to expedite computations. The attentiveness value is chosen as a criterion to identify the \textit{top-k} attentive tokens and fuse the rest. Evo-ViT \cite{evo-vit} goes further by updating and reintegrating the fused token into the network through a slow-fast evolution mechanism, preserving information more effectively.

Whereas the aforementioned soft pruning maintains spatial structure by preserving compressed token information, hard pruning methods adopt a more aggressive stance by completely removing tokens. ATS \cite{fayyaz2022ats} prunes tokens by scoring their importance using attention from the classification token and sampling them via inverse transform sampling. It adaptively selects a variable number of tokens per image, is parameter-free, and works with pre-trained models without retraining. CP-ViT \cite{CP-ViT} dynamically prunes uninformative patches and heads using cumulative attention-based scores computed across layers. AdaViT \cite{9879366} introduces a lightweight decision network integrated into each Transformer block, jointly optimized with the backbone. At inference time, it outputs binary decisions to selectively retain tokens, activate self-attention heads, or skip entire blocks, enabling dynamic and input-dependent computation. DynamicViT \cite{rao2021dynamicvit} also incorporates lightweight prediction modules at multiple layers to progressively estimate token importance and discard less informative ones. Zero-TPrune \cite{Zero-Tprune} applies a two-stage, zero-shot pruning process. It first ranks token importance using attention-based PageRank, then removes redundancy by merging similar tokens. Unlike AdaViT or DynamicViT, it requires no training or architectural modification, aligning more closely with ATS in its plug-and-play nature.

\subsection{Token merging}
Merging reduces the number of tokens by combining them into more informative, aggregated representations, while preserving key information. This can be done based on criteria like spatial proximity, semantic similarity, or predictive contribution. A common approach involves a hybrid of spatial and feature aggregation: spatial aggregation merges tokens from adjacent regions, while feature aggregation combines tokens with similar semantic representations. DPC-KNN \cite{zeng2022DPKNN} identifies clusters by estimating local token densities and merging those with minimal distance to high density points. TCFormer \cite{zeng2024tcformer} merges tokens from different locations through progressive clustering, generating new tokens with flexible shapes and sizes. AiluRus\cite{NEURIPS2023_62c9aa4d} reduces token count in ViTs via spatial-aware merging based on Density Peaks Clustering (DPC). Tokens are merged by selecting cluster centers using a score combining feature-space density and spatial distance. Non-center tokens are assigned to their nearest center. A reweighting mechanism adjusts attention to account for merged token groups. Token Pooling \cite{marin2023kmedoids} employs hard clustering by minimizing intra-cluster distances, using attention from the CLS token to initialize cluster centers. Following each transformer block, it identifies a subset of tokens that best approximates the underlying continuous signal, thereby capturing redundant features. ToMe \cite{bolya2022tome} computes token similarity using cosine similarity between attention keys, then merges the most similar token pairs using a bipartite matching algorithm. The merging is done via a weighted average of their features. ALGM  \cite{norouzi2024algm} performs token merging in a two-stage process. It first merges locally similar tokens in early layers, then globally merges semantically similar tokens mid-network, using cosine similarity and a ToMe-inspired strategy. LoTM \cite{haroun2024leveraging} introduces a local constraint by merging only pairs of horizontally adjacent tokens based on cosine similarity. DHTM \cite{haroun2025dynamic} extends the previous approach by considering all tokens as potential references and selectively merges only the most similar neighboring tokens in each Transformer layer. Unlike prior methods that rely on intermediate ViT features or fixed merging heuristics, DTEM \cite{lee2024dtem} learns a dedicated token embedding solely for merging. This decoupled embedding enables a soft, differentiable merging process during training and efficient hard merging at inference improving both flexibility and performance across tasks.



\subsubsection{Hybrid token reduction}
Determining whether to discard or merge tokens involves nuanced trade-offs, raising the issue of which strategy yields better performance for a particular task. Recent developments have introduced hybrid approaches that unify token merging and discarding within a single framework to further improve the efficiency of ViT. However, integrating both techniques introduces additional design considerations, particularly in determining when and how to apply each mechanism throughout the network. In this context, LTMP \cite{bonnaerens2023learned}introduces, into every Transformer block, threshold-based masking between MSA and FFN blocks to decide whether to keep, merge, or drop individual tokens. In ToFu \cite{kim2024tokenfusion}, the BSM algorithm plays a central role. Given a group of similar tokens, three token reduction strategies are proposed: tokens can either be fused using average merging, merged with MLERP (Norm-Preserving Average), or discarded. Token pruning strategy varies with layer depth: early layers apply discarding, while later layers favor merging. Both LTMP and ToFu adapt token merging from ToMe. PPT \cite{wu2024ppttokenpruningpooling} is based on the per-layer, per-instance variance of token importance scores. High variance favors pruning, while low variance favors merging. The authors observe that the variance of token importance scores increases with model depth, making token importance more distinguishable in deeper layers. Consequently, token pruning is more effective in deeper layers, while token merging is preferable in shallower layers; a finding that contrasts with the observations from ToFu. DiffRate \cite{chen2023diffrate} treats token reduction as a learnable optimization problem, allowing each layer to adjust its compression rate dynamically. Rather than handcrafting which layers should prune or merge tokens, DiffRate treats the compression rates as learnable parameters per layer. These are optimized during training through gradient descent, thanks to a module called the Differentiable Discrete Proxy (DDP). In practice, both token pruning and merging are applied in every transformer layer, but the proportion of each is learned in a differentiable manner. The pruning mechanism in UCC \cite{CHEN2025128747} is based on a hybrid importance score that combines both spatial and spectral information. At each Transformer block, tokens with low importance scores are pruned. However, instead of discarding them, UCC merges pruned tokens into the retained ones using a combination of cosine similarity and frequency-aware weighting, thereby maintaining the contextual integrity of the input. PM-ViT \cite{PM-ViT} proposes layer-wise compression strategy. This approach uses a learnable merge matrix to fuse less important tokens into aggregated representations and a reconstruct matrix to restore token dimensions after the transformer block. Token importance is estimated during training through a gradient-weighted attention scoring mechanism, which avoids extra computation at inference time. Tokens are categorized into three groups: high-importance tokens are preserved, medium-importance tokens are merged, and low-importance tokens are pruned. Shortcut connections are used to reintroduce pruned tokens, ensuring minimal information loss. 

\subsection{Token Reduction for Dense Tasks}
Most existing token reduction techniques have been primarily evaluated on image classification or generative tasks such as diffusion, with their applicability to dense prediction tasks remaining relatively underexplored. Token discarding methods, in particular, often involve the permanent removal of tokens that are deemed uninformative for the final prediction. This is feasible in classification settings due to the architectural design of ViTs, where the output is derived solely from the class token, which is always retained. However, in dense prediction tasks such as semantic segmentation, this strategy is not viable, as accurate pixel-wise predictions require preserving information from all spatial tokens. Consequently, more nuanced token reduction strategies like token merging or soft pruning are necessary to maintain spatial fidelity while reducing computational overhead. Consequently, only a few of the aforementioned token reduction methods are suitable for dense prediction tasks. Several approaches specifically developed for segmentation leverage merging-based mechanisms, whether at the patch \cite{lu2023cts, havtorn_msvit_2023} or token-level  \cite{norouzi2024algm}. ALGM extends ToMe’s global merging with segmentation-aware local merging and adaptive control, making it effective for dense prediction tasks. STViT \cite{huang2022stvit} and Ailurus \cite{NEURIPS2023_62c9aa4d} have been validated across various dense prediction tasks, including object detection, instance segmentation, and semantic segmentation. TCFormer \cite{zeng2022not, zeng2024tcformer} is presented as a general-purpose method applicable to various vision tasks, including object detection and semantic segmentation. However, its main limitation lies in the quadratic computational complexity of the KNN-DPC algorithm with respect to the number of tokens, which hampers its efficiency at high input resolutions. Among token-level pruning strategies, soft pruning is generally preferred, as it allows for more flexible token selection and gradual reduction without hard elimination. In particular, approaches that incorporate early stopping mechanisms appear especially well suited. In this direction, methods such as DToP \cite{Tang23DToP}, Paumer \cite{courdier2023paumerpatchpausingtransformer}, and DoViT  \cite{LiuWACV} have been specifically designed for semantic segmentation. SViT \cite{10483924} validates its approach on object detection and instance segmentation benchmarks. In contrast, DynamicViT \cite{rao2021dynamicvit} adopts hard token pruning, and its extended version \cite{10091227} also proves the method's effectiveness for object detection and instance segmentation. Hybrid token reduction methods, such as ToFu \cite{kim2024tokenfusion}, have shown promising results on image generation tasks. PM-ViT \cite{PM-ViT}, on the other hand, demonstrates its approach on image classification and semantic segmentation.

\section{Methodology}\label{STEPpipeline}

In this work, we introduce a novel token reduction strategy designed to improve the efficiency of ViTs. Our approach, named STEP (SuperToken and Early-Pruning), integrates two complementary techniques: supertoken generation and early token pruning. A supertoken is a compact representation derived from aggregating multiple spatially adjacent and semantically similar image patches into a single superpatch. The STEP mechanism, integrated into the vanilla ViT architecture, effectively reduces sequence length while preserving the essential spatial and semantic structure of the image, resulting in a more efficient yet accurate segmentation pipeline. This is achieved through dynamic adjustment of patch merging rates and token halting.

\subsection{Motivation}

Token-level pruning strategies applied deeper in the network often rely on intermediate attention scores or learned token importance, requiring additional computation and training complexity. These methods also typically maintain the full input sequence during early layers. From our point of view, reducing token count at the patch-level provides several practical and architectural benefits compared to token-level pruning. Since patches constitute the input units for the ViT, eliminating redundant ones at this stage directly shortens the input sequence. This leads to immediate reductions in computational cost and memory usage across all subsequent layers. The impact is particularly significant in the case of high-resolution semantic segmentation, where the initial number of tokens can be extremely high. Moreover, patch-level reduction is inherently more interpretable and compatible with pre-trained models. 

\begin{figure}[ht]
\centering
\includegraphics[width=0.6\textwidth]{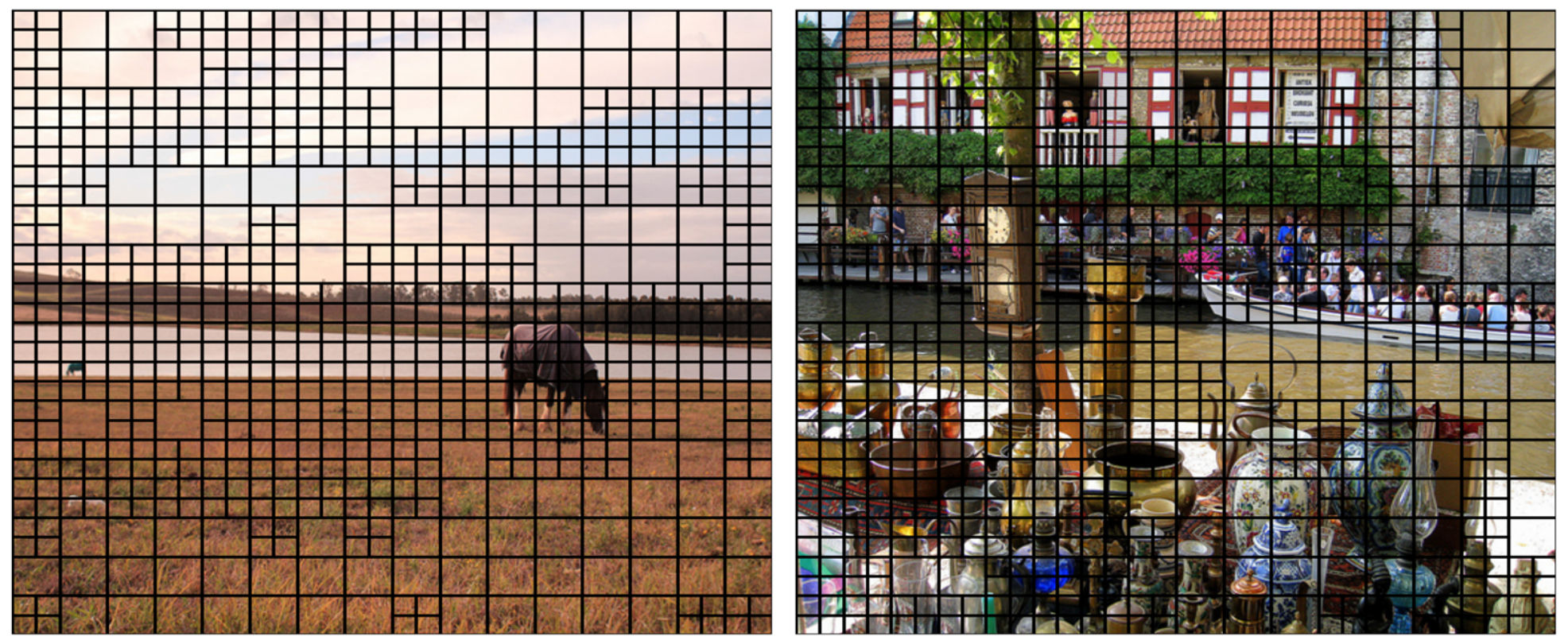}
\caption{Example of failures of CTS due to the top-K merging strategy. Left: Too few merges on simple images; Right: Too many merges on complex images.}
\label{CTS_issues}
\end{figure}

However, existing approaches have certain limitations. CTS \cite{lu2023cts}, for example, fixes the number (K=103) and size ($2\times2$) of superpatches as hyperparameters. This can be problematic for complex images, as it may lead to the merging of patches that should remain separate. Conversely, for images with homogeneous content, the merging rate can often be suboptimal (see Figure~\ref{CTS_issues}). MSViT \cite{havtorn_msvit_2023} addresses one of the limitations of CTS by dynamically adapting the number of merged patches. Nevertheless, the highest resolution patch remains limited to $2\times2$. Quadformer employs three grouping sizes namely $8\times8$, $4\times4$, and $2\times2$ patches in its mixed-resolution tokenization scheme. A key limitation of this approach is the increased inference-time overhead observed in small models, especially ViT-Small. Although the saliency scorer is lightweight in terms of parameters, its execution is relatively slow compared to the fast inference of compact models. This motivates our focus on content-adaptive patch-level pruning, which dynamically adjusts supertoken resolution and number based on local semantic homogeneity. We retain a regular grid structure after patch merging to simplify positional embedding interpolation and maintain compatibility with standard ViT architectures. 

Moreover, we believe, as demonstrated by soft pruning approaches, that tokens vary in difficulty, and that simpler tokens may be predicted earlier, eliminating the need for a complete forward pass through the entire network. Once a sufficient confidence level is reached, their further processing can be halted.  In segmentation tasks, this idea becomes even more appealing, since tokens cannot be entirely removed due to the requirement for per-token predictions. We therefore consider this method to be complementary to input sequence length reduction, as it enables a progressive shortening of the set of tokens processed as the network deepens. Such a hybrid approach not only reduces computation but also allows ViTs to better allocate attention and processing power to semantically rich regions, making them more suitable for high-resolution semantic segmentation.

\subsection{Overview of the STEP}
\begin{figure}[htbp]
    \centering
    \includegraphics[width=\textwidth]{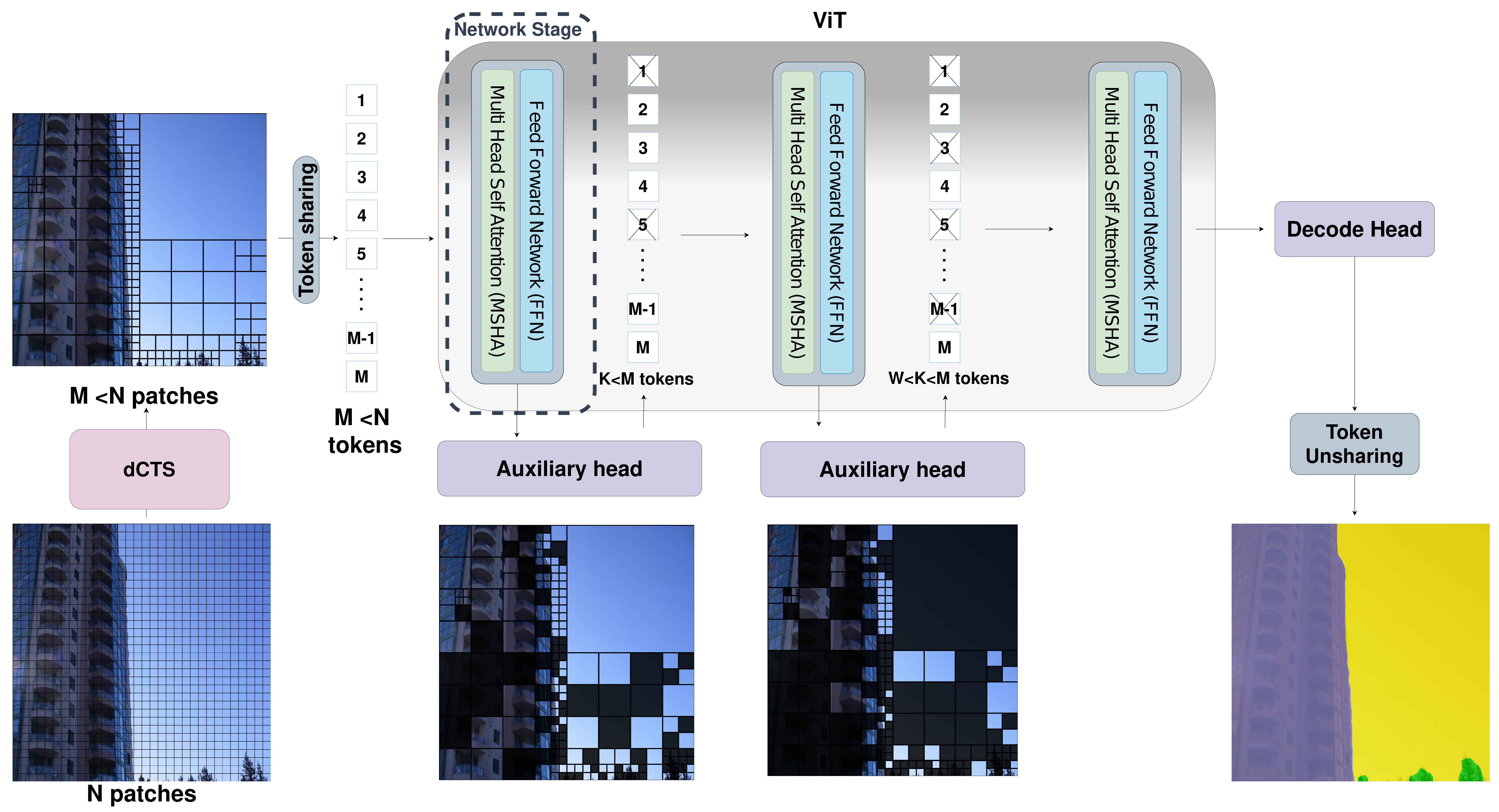}   \caption{STEP overview. The image is first divided into fixed-size grid patches. The dCTS policy network then predicts which groups of patches can be merged into superpatches of varying resolutions, which are subsequently transformed into supertokens. Following the DToP approach, the ViT encoder blocks are organized into \(S\) stages. This multi-stage structure, equipped with auxiliary decoders, dynamically masks high-confidence tokens (represented as black squares), while the remaining tokens are propagated through the subsequent layers. The final decoder head combines predictions from all stages to generate the final output. Figure inspired from \cite{proust2025step}.}
    \label{fig:STEP}
    
\end{figure}

STEP is a hybrid token-reduction approach that operates on two levels: it first merges patches at the local level, then performs additional token pruning at selected stages of the network (Figure~\ref{fig:STEP}). The process begins by dividing the image into a uniform grid of superpatches, following the standard procedure used in vanilla Transformers (in our case, 16$\times$16 pixel patches). Next, a module called dCTS performs token merging based on similarity, resulting in a grid of superpatches with non-uniform sizes. The token-sharing module transforms the created superpatches into supertokens. Superpatches are resized to the standard $16\times16$ pixel resolution using bilinear interpolation and projected into the embedding space in the same way as regular patches. The latter is performed by applying a linear embedding function $f_{\text{embed}}$, which maps the superpatches into their corresponding token representations:

\begin{equation}
Z = f_{\text{embed}}(P')\label{eq1}
\end{equation}

where $P'$ represents the set of superpatches, and $f_{\text{embed}}$ is the linear embedding function that generates the supertokens $Z$.
The transformer-based ViT models
process the resulting supertokens and produce the final
output through per-token predictions.
An early exit strategy is also implemented through an auxiliary decoding head within encoder blocks, which are divided into $S$ stages. This allows tokens that are confidently predicted by the model in the early layers to be halted, thereby reducing overall computational costs without compromising segmentation accuracy. Only the most challenging tokens continue to propagate through the deeper layers of the transformer.

\subsubsection{Semantic-aware patch aggregation} \label{dCTS}
We propose a flexible and content-adaptive strategy, referred to as dynamic CTS (dCTS), inspired by CTS but designed to more effectively address the inherent complexity and variability of image contents. This merging step is guided by a lightweight class-agnostic policy network built upon the EfficientNetLite0 architecture \cite{Tan2019EfficientNetRM}, pre-trained on ImageNet-1K \cite{ILSVRC15}. For each image, groups of adjacent patches are considered and a similarity score is computed to assess whether the group likely belongs to a single semantic class. Given any window of \( n \) neighboring patches \( \mathcal{W} = \{ \mathbf{p}_1, \mathbf{p}_2, \ldots, \mathbf{p}_n \} \), the policy network predicts a similarity score \( \mathbf{S} \) as follows:

\begin{equation}
\mathbf{S} = \sigma \left( \mathbf{W}_p^\top \left( \mathcal{W} \right) \right)
\label{eq2}
\end{equation}

where \( \mathbf{W}_p \) is the learned weight matrix of the policy network and \( \sigma \) denotes the sigmoid activation function. Fusion is performed using a threshold-based approach: if the similarity score exceeds a predefined threshold \( \tau \), the patches are merged into a superpatch:

\begin{equation}
\mathbf{p}^{\text{sp}} = \text{concat}(\mathbf{p}_1, \mathbf{p}_2, \ldots, \mathbf{p}_n)
\label{eq:eq3}
\end{equation}

\begin{figure}[ht]
\centering
\includegraphics[width=\textwidth]{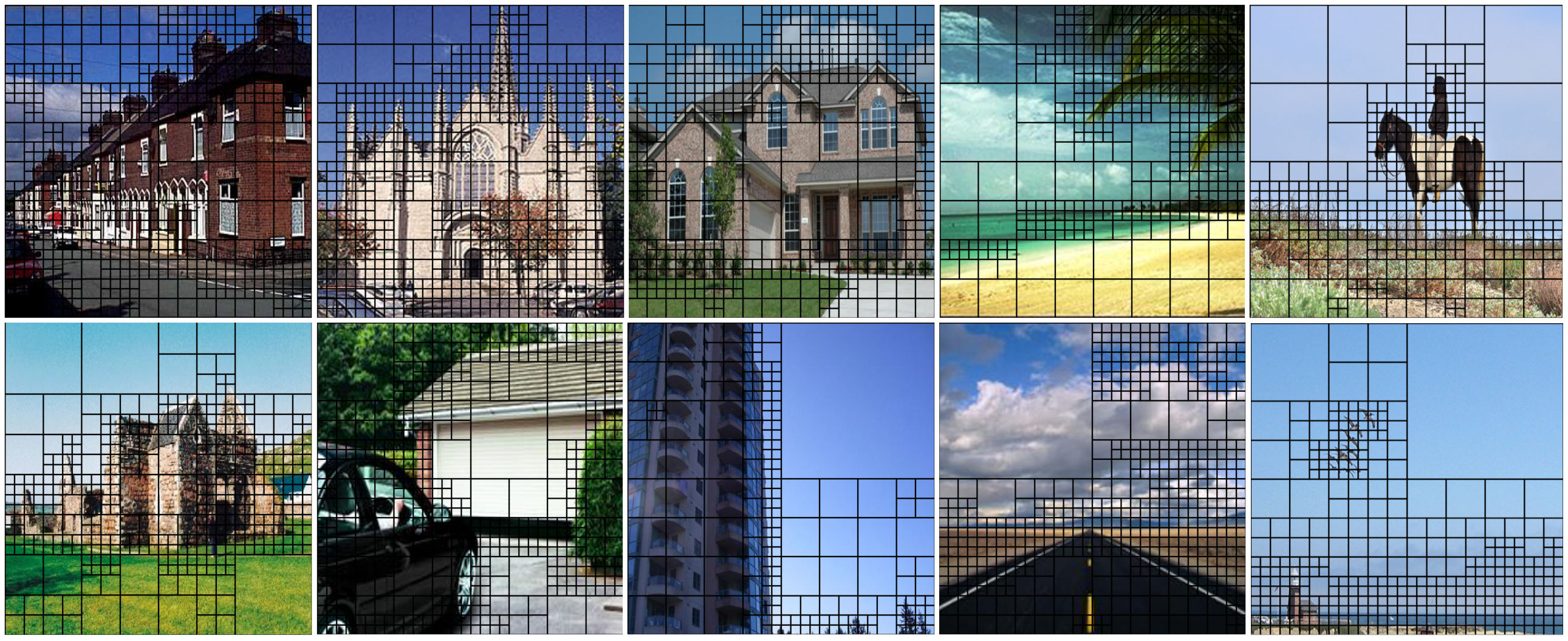}
\caption{Resolution-aware splitting on superpatches using dCTS (from $2\times2$ to $16\times16$).}
\label{dCTS_ADE20k}
\end{figure}

In practice, the policy network processes the input image after it has been divided into uniform patches. For each group, it produces a similarity score \( \mathbf{S} \), a continuous value in the range \([0, 1]\), which is interpreted as the probability that the group is homogeneous. Rather than predicting the exact class, the network leverages this probability and applies a predefined threshold \( \tau \) to categorize each group into one of two classes: (i) likely belonging to a single semantic class or (ii) likely heterogeneous. This probabilistic interpretation supports a flexible, threshold-based decision mechanism for patch merging. Fusion proceeds in a coarse-to-fine manner, starting with the largest window sizes ($16\times16$ patches) and progressively evaluating smaller windows ($8\times8$, $4\times4$, and $2\times2$). This hierarchical order ensures that there are no conflicts between nested groups \textit{, i.e.}, it prevents the merging of a smaller $2\times2$ patch group that is already part of a larger region deemed mergeable. Figure~\ref{dCTS_ADE20k} presents illustrative results of the merging process using our dCTS method. These examples showcase the ability of dCTS to adaptively merge patches in homogeneous regions (e.g., background or sky) while maintaining higher spatial resolution in semantically complex areas such as object boundaries or textured regions.

\subsubsection{Early-pruning mechanism} \label{DToP}

We incorporate a state-of-the-art early exiting mechanism inspired by DToP into our pipeline. The core principle behind DToP is to identify easy-to-predict tokens at intermediate layers and exclude them from further processing. To achieve this, the model is structured into \( M \) sequential stages. After a fixed number of attention blocks, an auxiliary head computes a confidence score \( c_i^{(m)} \) for each token \( \mathbf{z}_i \). Tokens whose confidence exceeds a predefined threshold \( \tau \) are considered high confidence and are masked out, that is, removed from subsequent computation.  Low-confidence tokens \( \mathbf{Z}^{(m+1)} \) continue to propagate through the subsequent encoder layers :
\begin{equation}
\label{eq:eq4}
\mathbf{Z}^{(m+1)} = \left\{ \mathbf{z}_i \;\middle|\; c_i^{(m)} < \tau \right\}
\end{equation}

Similar to DToP, our implementation employs auxiliary heads that adopt the attention-to-mask (ATM) module \cite{zhang2022segvit}. These heads are architecturally identical to the final decoder head, ensuring consistent behavior across all stages. The effectiveness of such an early-pruning mechanism heavily depends on the proper placement of the auxiliary heads. If placed too early in the network, the model may fail to generate reliable predictions, as the representations are not yet sufficiently informative. Conversely, if the auxiliary heads are positioned too late in the network, most of the computational cost has already been incurred by the time pruning occurs. As a result, the potential savings in inference time and FLOPs are significantly reduced, defeating the main purpose of early exiting. The authors of DToP introduce auxiliary heads at specific layers, namely the 6th and 8th for ViT-Base, and the 8th and 16th for ViT-Large. Although this configuration yields a reasonable trade-off between computational cost and segmentation accuracy, it remains largely empirical and lacks a principled justification. The exploration of pruning positions is limited to a small set of static configurations, and the impact of pruning positions on inference time is not explicitly discussed. We argue that further investigations are needed to establish more generalizable and adaptive guidelines for auxiliary head placement. This includes studying the internal evolution of token difficulty, exploring data- or budget-adaptive strategies, and considering the impact of auxiliary head placement on real-time inference efficiency.

\section{Experiments}\label{sec2}
This section presents a detailed evaluation of our STEP mechanism on widely used benchmarks, focusing on both predictive accuracy and computational efficiency. The evaluation begins with a description of the main architectural and hyperparameter choices involved in the design of our STEP method. In particular, we analyze the impact of threshold parameters that control the semantic-aware patch merging via dCTS (Section~\ref{dCTS_config}). We also investigate the placement and configuration of early-exit branches, with a focus on their number and depth within the transformer architecture (Section~\ref{DToP_config}). To comprehensively assess STEP performance and isolate the effect of each component, mean Intersection over Union (mIoU) is used to evaluate segmentation accuracy, while GFLOPs (giga floating-point operations) provide an estimate of the model’s computational complexity. GFLOPs are computed using the fvcore package\footnote{https://github.com/facebookresearch/fvcore}, ensuring consistent measurement across all configurations. These metrics are reported for both standard-resolution and high-resolution settings (Section~\ref{Results}). 

\subsection{Experimental Setup}
We integrate STEP \cite{zhang2022segvit} into the SegViT semantic segmentation framework. All experiments are conducted using MMSegmentation\footnote{https://github.com/open-mmlab/mmsegmentation}\cite{mmseg2020}, an open-source PyTorch-based library that facilitates flexible backbone integration. We evaluate our approach using both ViT-Base and ViT-Large models. ViT-Base includes 12 transformer encoder layers, a hidden size of 768, and 12 attention heads, while ViT-Large consists of 24 layers, a 1024-dimensional hidden state, and 16 attention heads. In both cases, input images are first divided into a non-overlapping $16\times16$ pixel grid of patches.
Experiments are conducted on three widely used semantic segmentation benchmarks: COCOStuff10k \cite{8578230}, which includes a wide variety of objects in complex, real-world scenes, ADE20K \cite{zhou2017scene}, a comprehensive dataset for scene parsing, and Cityscapes \cite{Cordts2016Cityscapes}, which focuses on urban street scenes with high-quality pixel-level annotations. The standard evaluation is conducted using fixed input resolutions, namely $512\times512$ in accordance with commonly adopted benchmarking protocols. To evaluate scalability under high-resolution conditions, additional experiments are conducted on the Cityscapes dataset, which offers images with a consistent resolution of $2048\times1024$. The DToP confidence threshold is set to 0.95 for COCOStuff10k, and 0.9 for ADE20K and Cityscapes. Optimization is performed using AdamW with an initial learning rate of 6e-5, a weight decay of 0.01, and a cosine learning rate schedule. Training follows the standard mmseg configuration. Models are trained for 160K iterations on ADE20K, 80K iterations on COCOStuff10k, and 90K for Cityscapes with a batch size of 4. Data augmentation includes random horizontal flipping, resizing with a scale ratio between 0.5 and 2.0, and random cropping. We acknowledge that the chosen parameters may not be optimal for achieving the highest possible performance (e.g., mIoU). However, our primary objective is not to maximize accuracy, but rather to demonstrate the efficiency gains enabled by our token reduction approach.

\subsection{dCTS Under Varying Thresholds}\label{dCTS_config}

\begin{figure}[htbp]
    \centering
    \begin{subfigure}[b]{0.48\textwidth}
        \centering
        \includegraphics[width=\textwidth]{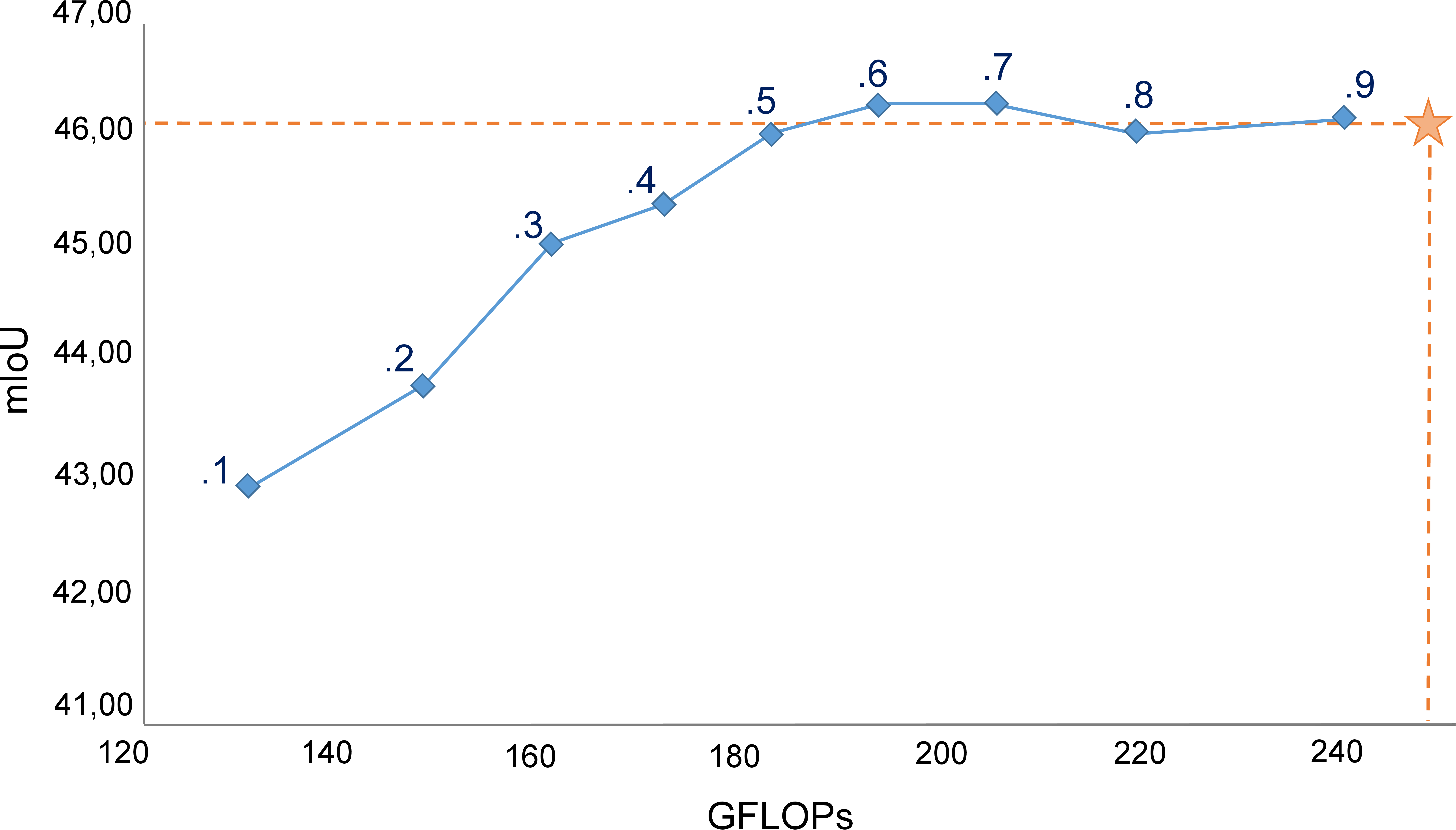}
        \caption{Cocostuff10k}
    \end{subfigure}
    \hfill
    \begin{subfigure}[b]{0.48\textwidth}
        \centering
        \includegraphics[width=\textwidth]{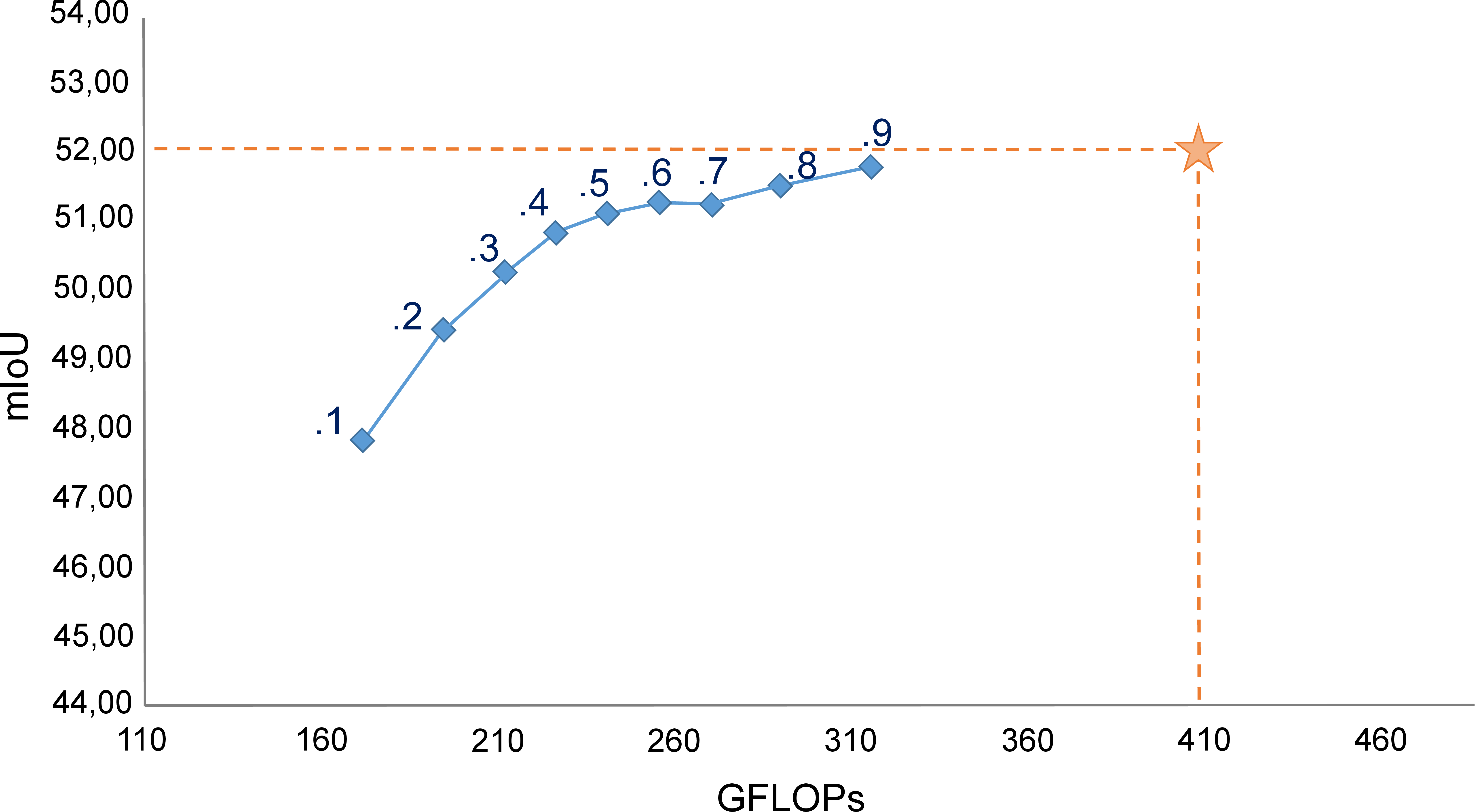}
        \caption{ADE20k}
    \end{subfigure}
    \caption{Tuning the merging threshold \( \tau \) hyperparameter influences both segmentation accuracy and computational cost when employing the ViT-Large backbone.
    The blue curve illustrates the effect of varying \( \tau \) when merging only $2\times2$ patch groups. The orange star indicates the performance of CTS with a fixed number of merged patches.}
    \label{fig:Thres2x2}
\end{figure}

We conduct a series of experiments to determine the optimal merging threshold \( \tau \) for various superpatch sizes in our dCTS approach. In this process, we assess model performance in terms of mIoU and GFLOPs, using ViT-Large as the backbone and two different datasets, with standard image resolutions typically used for segmentation tasks. This enables us to identify the best trade-off between computational efficiency and segmentation accuracy for each superpatch size. For example, when merging only $2\times2$ patch groups, we find that setting the threshold to \( \tau = 0.4 \) achieves the best trade-off between accuracy and computational cost. This configuration leads to a modest accuracy drop of approximately 1\%, while reducing computational complexity by at least 30\%. This trend is consistently observed across both datasets (see Figure~\ref{fig:Thres2x2}).

In our dCTS approach, we apply the same principle by assigning a distinct threshold value \( \tau \) to each patch group size. Specifically, we set a high threshold of 0.9 for larger patch groups, while lower values are used for smaller ones, starting from \( \tau = 0.4 \) for the smallest $2\times2$ groups. This strategy is motivated by the need to prevent errors when forming large superpatches, as incorrect merges at this scale can significantly degrade the quality of the final segmentation. Table~\ref{dCTS_Coco_VitL} summarizes the results obtained for several
threshold  \( \tau \) configurations. From this, we determine the optimal combination to be \( \tau \)-4999 or \( \tau \)-6899 for the $2\times2$, $4\times4$, $8\times8$, and $16\times16$ superpatch sizes, respectively. Compared to the CTS, the first configuration allows no loss in segmentation accuracy while reducing computational complexity by 27\%. The second is less strict on segmentation quality, allowing a potential 1\% loss in mIoU, but reducing complexity by 36\%. 

\begin{table}[htbp]
\caption{Performance of the dCTS method on the COCOStuff10k dataset using size-dependent merging thresholds \( \tau \) for different superpatch sizes $2\times2$, $4\times4$, $8\times8$, and $16\times16$.}

\label{dCTS_Coco_VitL}
\centering
\begin{tabular}{lccccccccc}
\toprule
& \multicolumn{8}{c}{Threshold \( \tau \) } \\
\cmidrule(lr){3-10}
Metric & CTS & .6.9.9.9 & .6.8.9.9 & .4.9.9.9 & .4.8.9.9 & .4.7.9.9 & .4.6.9.9 & .4.5.9.9 & .4.4.9.9 \\
\midrule
mIoU & 46.1 & 45.9 & 46.0 & 45.3 & 44.8 & 43.9 & 44.1 & 43.7 & 43.7 \\
\hline
\hline
GFLOPs & 248 & 189 & 181 & 159 & 156 & 153  & 151 & 149 & 147 \\
\bottomrule
\end{tabular}
\end{table}

As shown in Table~\ref{dCTS_stats_merged}, on high-resolution images, an average of 2988 patches (out of 4096) are merged using dCTS, representing an increase compared to the 412 patches merged with CTS. It can be also observed that the merging of larger neighboring patch groups such as $8\times8$ and $16\times16$ remains relatively rare. This is consistent with the nature of Cityscapes, which mostly contains visually complex scenes with multiple objects and diverse textures, where large homogeneous regions are relatively uncommon. Nonetheless, dCTS achieves on average a $2.5\times$ reduction in the number of patches on high-resolution images. This trend is also confirmed by experiments on other standard-resolution datasets \cite{proust2025step}, where the merging approach enables up to a 6× token reduction for highly homogeneous content, and up to 3× for more complex scenes, compared to standard fixed-grid slicing.
 
\begin{table}[htbp]
\caption{Statistical insights into token pruning via STEP on the Cityscapes dataset for different input resolutions.}
\label{dCTS_stats_merged}
\centering
\begin{tabular}{l ccccc ccccc}
\toprule
& \multicolumn{5}{c}{Input $512\times512$} & \multicolumn{5}{c}{Input $1024\times1024$} \\
\cmidrule(lr){2-6} \cmidrule(lr){7-11}
Metric & \multicolumn{5}{c}{Superpatch resolution} & \multicolumn{5}{c}{Superpatch resolution} \\
\cmidrule(lr){2-6} \cmidrule(lr){7-11}
& $1\times1$ & $2\times2$ & $4\times4$ & $8\times8$ & $16\times16$ & $1\times1$ & $2\times2$ & $4\times4$ & $8\times8$ & $16\times16$ \\
\midrule
Mean    & 640 & 52 & 9  & 0.3 & 0 & 1108 & 452 & 50 & 10  & 0.8  \\
Maximum & 816 & 84 & 20 & 3   & 0 & 2212 & 656 & 32 & 21  & 5    \\
Minimum & 424 & 24 & 0  & 0   & 0 & 632  & 246 & 2  & 0   & 0    \\
\bottomrule
\end{tabular}
\end{table}

\subsection{Prune Smart: Where Should Tokens Exit?} \label{DToP_config}
\begin{figure}[htbp]
  \centering
  \includegraphics[width=0.5\textwidth]{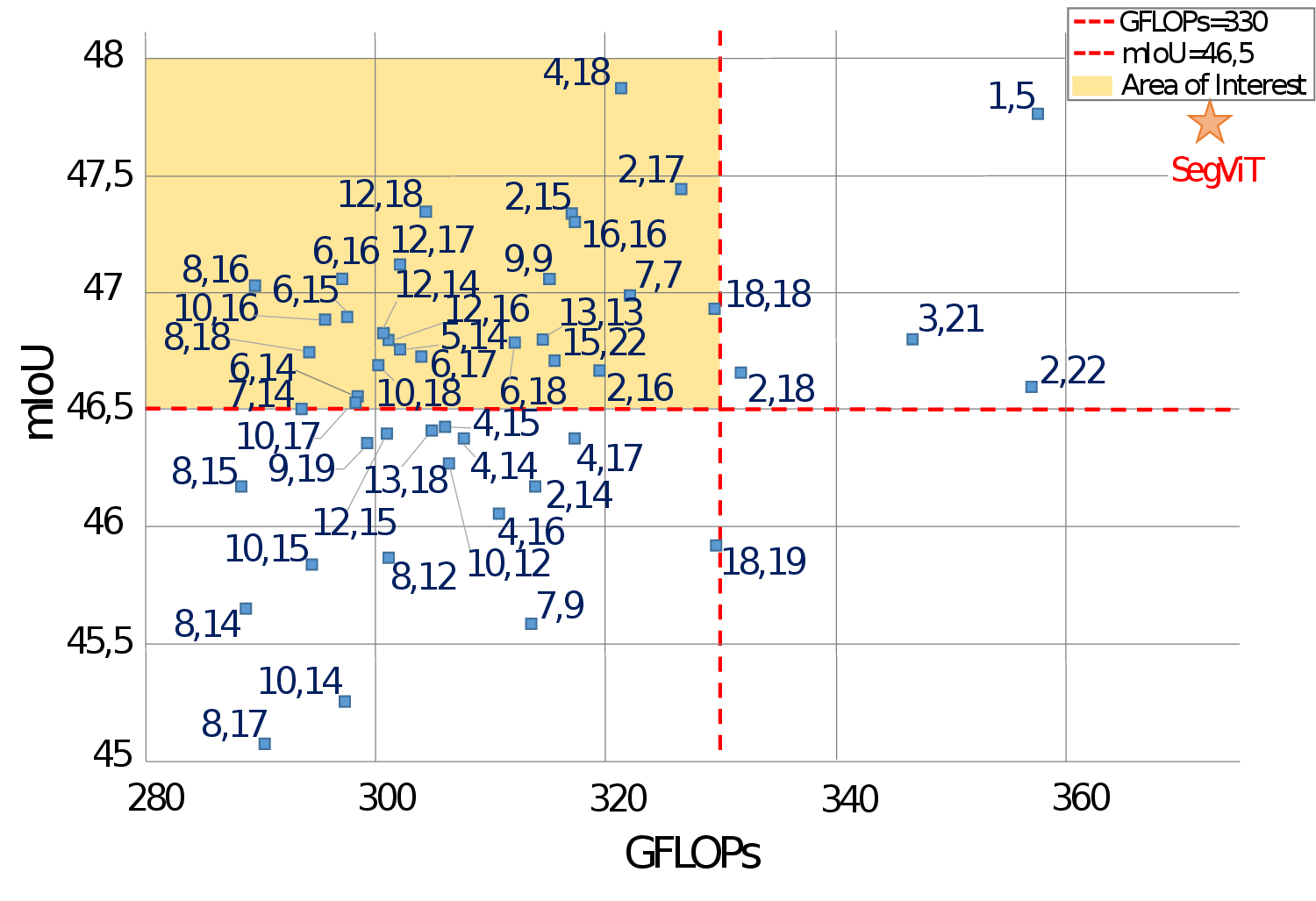}
  \caption{Pruning head configuration analysis on the COCOStuff10k dataset. The numbered markers indicate the positions of the auxiliary heads, while the  star corresponds to the performance of the baseline SegViT model without pruning. The plot illustrates the trade-off between segmentation accuracy (mIoU) and computational complexity (GFLOPs). Configurations within the yellow rectangle are selected for further analysis, as they yield at least a 10\% reduction. Figure from \cite{proust2025step}.}
  \label{fig:FlOPSvsMiOU}
 \end{figure}

\begin{figure}[htbp]
    \centering
    \begin{subfigure}[b]{0.48\textwidth}
        \centering
        \includegraphics[width=\textwidth]{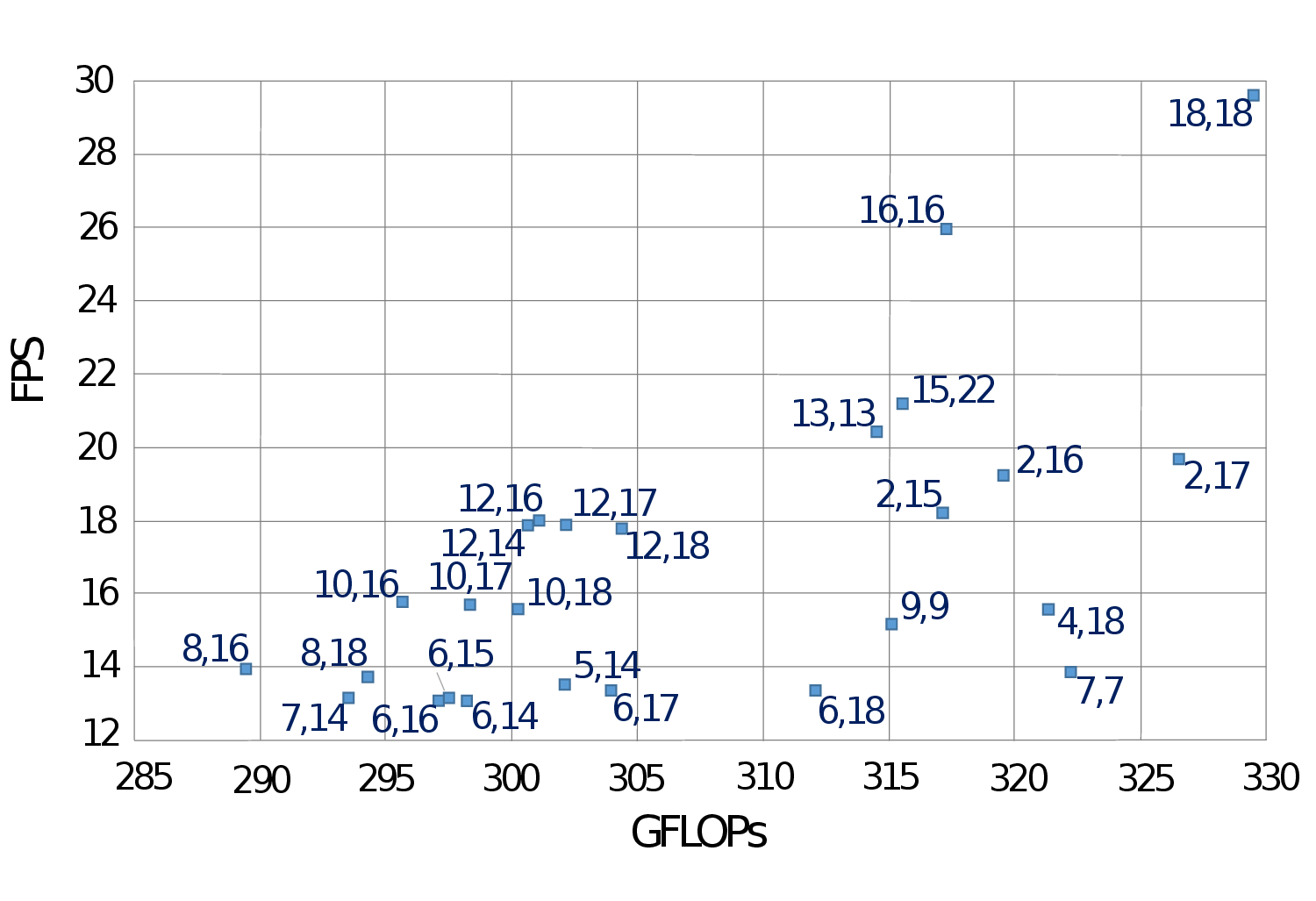}
        \caption{Trade-off between throughput and computational cost.}
        \label{fig:FLOPs_FPS}
    \end{subfigure}
    \hfill
    \begin{subfigure}[b]{0.48\textwidth}
        \centering
        \includegraphics[width=\textwidth]{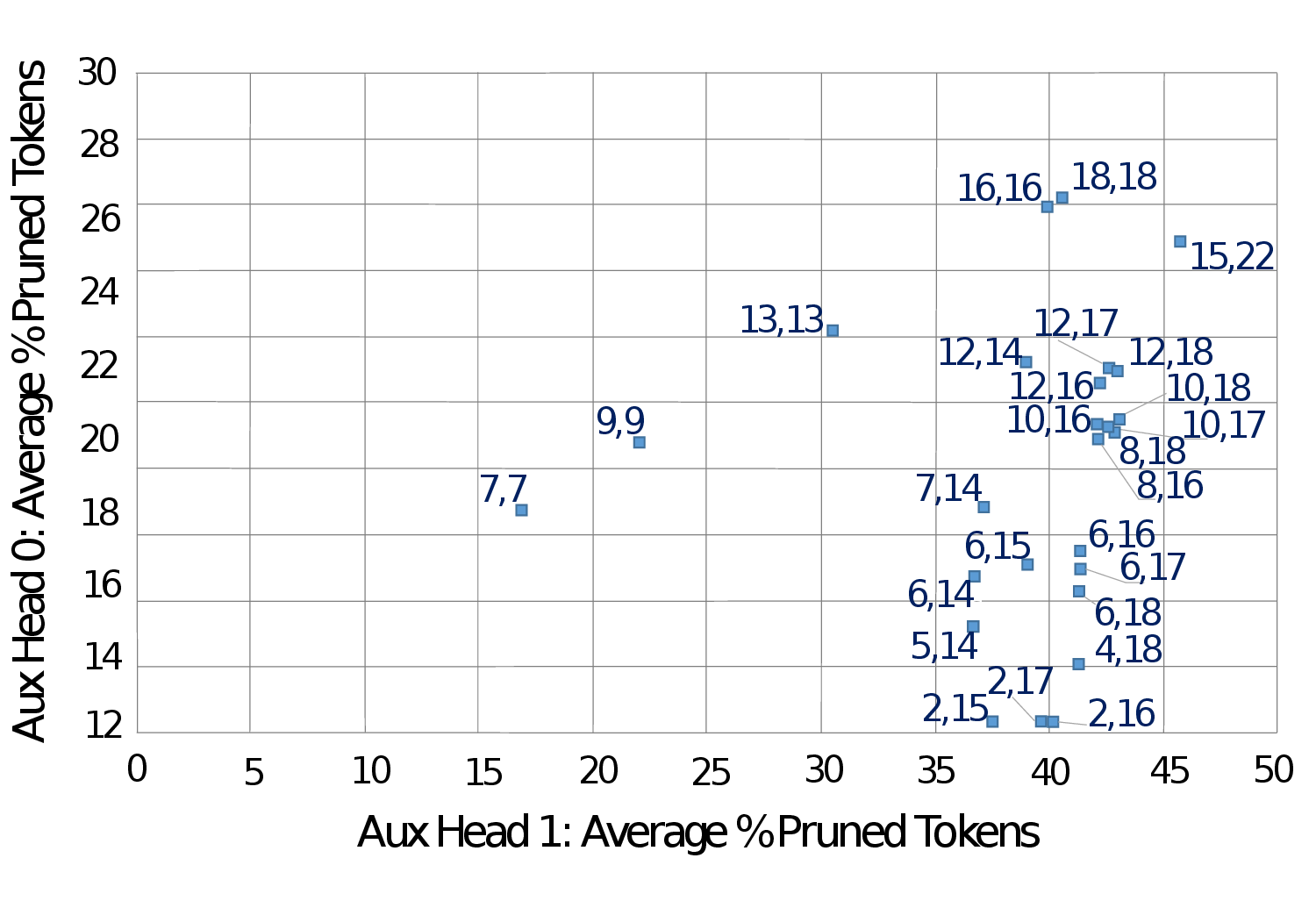}
        \caption{Average percentage of pruned tokens per configuration.}
        \label{fig:HowManyPruned}
    \end{subfigure}
    \caption{Exploration of the pruning head configuration on the COCOStuff10k dataset. Figure from \cite{proust2025step}.}
\end{figure}

To determine the optimal placement of the auxiliary head, we conduct an ablation study of the early-exit mechanism based on DToP, using standard-resolution images commonly employed in semantic segmentation benchmarks. We choose to partition the large ViT backbone (24 encoder layers) into a maximum of three stages. For each configuration, we evaluate its impact on segmentation accuracy, computational complexity (Figure~\ref{fig:FlOPSvsMiOU}), inference time (Figure~\ref{fig:FLOPs_FPS}), and the percentage of pruned tokens (Figure~\ref{fig:HowManyPruned}), in order to identify the most effective positioning strategy.
The results clearly demonstrate that the number and placement of auxiliary heads directly impact computational complexity and inference speed. For instance, placing two auxiliary heads at the 8th and 16th layers achieves a 22\% reduction in GFLOPs (289 vs. 373), while maintaining segmentation accuracy comparable to the baseline SegViT model, which performs no token pruning. However, this gain in efficiency comes at the expense of throughput, with inference time increasing threefold compared to the unpruned baseline. In contrast, using a single auxiliary head placed deeper in the network (e.g., at the 16th or 18th layer) offers a more favorable trade-off. Although it slows inference, it still provides a significant reduction in computational cost. Figure~\ref{fig:HowManyPruned} further shows that with a single pruning head the percentage of pruned tokens increases linearly, reaching around 40\% on average. Remarkably, this level of pruning is comparable to what is achieved with two early-exit heads, regardless of their configuration. This suggests that a well-placed single auxiliary head can be nearly as effective as a more complex multi-head setup.

Identifying the optimal pruning configuration is a non-trivial and nuanced process. If the primary goal is to reduce computational complexity, our results indicate that splitting the large model (i.e., ViT-Large) into two stages and inserting auxiliary heads after the 8th and 16th layers yields the most effective token pruning. However, if inference speed is the main concern, a more suitable approach is to use only a single auxiliary head, positioned as early as the 16th layer, which balances token reduction with acceptable latency overhead. To adapt this strategy to smaller 12-layer architectures like ViT-Base, we interpolate our results and identify the 8th layer as the optimal position for deploying a single auxiliary head.

\subsection{Results and Discussion}\label{Results}
To enable a fair evaluation, we compare our token reduction mechanism against SegViT in its original form.  We also construct its pruned variants by applying state-of-the-art token reduction techniques. Specifically, the CTS method is used for patch merging, followed by soft token pruning using DToP. A combined configuration incorporating both techniques is also evaluated, as it represents a preliminary version of our STEP mechanism. Throughout this process, we adhered to the baseline configurations and parameters established
by the authors. We combine a fixed number of $2\times2$ patches for CTS, specifically merging 103 patches, and position the auxiliary heads at the 8th and 16th layers for DToP. In our STEP method, we apply the previously described threshold configuration for dCTS. We choose to divide the ViT-Large model into two and three stages, naming them STEP@[18] and STEP@[8,16], respectively. The values in brackets indicate the pruning heads positions. For experiments on ViT-Base, we adopt the configuration proposed by the original DToP, placing auxiliary pruning heads after the 6th and 8th encoder layers. In addition, we evaluate our own strategy by applying a single pruning head after the 8th layer, as a lighter alternative aiming for better inference efficiency.

Table~\ref{Lowresolution_STEP_ViTL} reports the performance of STEP integrated into ViT-L with standard low-resolution inputs. The results indicate that STEP achieves segmentation accuracy comparable to the baseline, with mIoU degradation remaining below 2.5\% across configurations. Moreover, it consistently yields a substantial reduction in computational complexity across different datasets. Introducing two auxiliary heads further amplifies this gain, achieving up to a 2.8× reduction in GFLOPs. However, this comes at the cost of significantly lower throughput. To ensure the robustness of our conclusions, we also replicate the experiments using ViT-Base as the backbone. The corresponding results are reported in Table~\ref{STEP_ViTB}. Our STEP method achieves up to a 2.5× reduction in computational cost compared to the SegViT baseline, while incurring an accuracy drop comparable to that observed with ViT-Large. We further assess the effect of patch fusion on performance using our dCTS\( \tau \)-6899 variant. Notably, it achieves the best trade-off by reaching 48.2 mIoU on ADE20k, which is identical to the baseline, while requiring only 73 GFLOPs and delivering a high inference speed of 98 FPS, nearly twice as fast.

\begin{table}[htbp]
\caption{Performance evaluation of our STEP mechanism integrated into ViT-Large.}\label{Lowresolution_STEP_ViTL}
\begin{tabular*}{\textwidth}{@{\extracolsep\fill}lcccccc}
\toprule%
& \multicolumn{3}{@{}c@{}}{ADE20k ($512\times512$)} & \multicolumn{3}{@{}c@{}}{COCOStuff10k ($512\times512$)} \\\cmidrule{2-4}\cmidrule{5-7}%
Method & mIoU$\nearrow$ & GFLOPs$\searrow$ & FPS$\nearrow$ & mIoU$\nearrow$ & GFLOPs$\searrow$ & FPS$\nearrow$ \\
\midrule
SegViT                 & 53.0 & 624 & 38 & 46.7 & 373 & 44.5\\
+CTS\footnotemark[1]   & 52.0 & 410 & 41 & 46.2 & 251 & 40 \\
+DToP\footnotemark[1]  & 52.3 & 465 &  6 & 46.6 & 290 & 15\\
\hline
\hline
+CTS\footnotemark[1] \& DToP\footnotemark[1]  & 51.2 & 334  & 12.5 & 45.4 & 210 & 17 \\
+STEP@[8,16]\( \tau \)-6899 & 51.2 & 224 & 14   & 46.0 & 173 & 18 \\
+STEP@[8,16]\( \tau \)-4999 & 50.8 & 334 & 15   & 45.3 & 150 & 20 \\
+STEP@[18]\( \tau \)-6899   & 51.7 & 395 & 22   & 46.0 & 201 & 30 \\
+STEP@[18]\( \tau \)-4999   & 50.4 & 261 & 26.5 & 45.1 & 177 & 29 \\
\botrule
\end{tabular*}
\footnotetext[1]{Default configuration from the original paper}
\end{table}

\begin{table}[htbp]
\caption{Performance evaluation of our STEP mechanism integrated with ViT-Base.}\label{STEP_ViTB}
\begin{tabular*}{\textwidth}{@{\extracolsep\fill}lcccccc}
\toprule%
& \multicolumn{3}{@{}c@{}}{ADE20k ($512\times512$)} & \multicolumn{3}{@{}c@{}}{Cityscapes ($512\times512$)} \\\cmidrule{2-4}\cmidrule{5-7}%
Method & mIoU$\nearrow$ & GFLOPs$\searrow$ & FPS$\nearrow$ & mIoU$\nearrow$ & GFLOPs$\searrow$ & FPS$\nearrow$ \\
\midrule
SegViT                      & 48.3 & 113 & 53 & 67.7 & 110 & 70 \\
+CTS\footnotemark[1]        & 47.8 & 75  & 40 & 67.6 & 73  & 56 \\
+DToP\footnotemark[1]       & 45.8 & 91  & 25 & 68.2 & 82  & 24 \\
\hline
\hline
+CTS\footnotemark[1] \& DToP\footnotemark[1]  & 46.3 & 62 & 25 & 67.5 & 59.5 & 27   \\
+dCTS \( \tau \)-6899       & 48.2 & 73 & 98 & 67.5 & 77 & 66 \\
+STEP@[6,8]\( \tau \)-6899  & 46.9 & 64 & 24 & 67.2 & 61 & 22 \\
+STEP@[6,8]\( \tau \)-4999  & 45.3 & 50 & 34 & 64.3 & 44 & 26 \\
+STEP@[8]\( \tau \)-6899    & 47.1 & 68 & 32 & 67.4 & 66 & 32 \\
+STEP@[8]\( \tau \)-4999    & 45.8 & 53 & 43 & 64.2 & 44 & 32 \\

\botrule
\end{tabular*}
\footnotetext[1]{Default configuration from the original paper}
\end{table}

\begin{table}[htbp]
\caption{Performance evaluation of our STEP mechanism integrated with ViT-Base.}\label{STEP_ViTB_Cityscapes_high}
\begin{tabular*}{\textwidth}{@{\extracolsep\fill}lcccccc}
\toprule%
& \multicolumn{3}{@{}c@{}}{Cityscapes ($768\times768$)} & \multicolumn{3}{@{}c@{}}{Cityscapes ($1024\times1024$)} \\\cmidrule{2-4}\cmidrule{5-7}%
Method & mIoU$\nearrow$ & GFLOPs$\searrow$ & FPS$\nearrow$ & mIoU$\nearrow$ & GFLOPs$\searrow$ & FPS$\nearrow$ \\
\midrule
SegViT                  & 73.7 & 301 & 65 & 75.2 & 670 & 24 \\
+CTS\footnotemark[1]    & 72.9 & 190 & 45 & 74.9 & 403 & 46 \\
+DToP\footnotemark[1]   & 73.5 & 198 & 22 & 75.0 & 430 & 16 \\
\hline
\hline
+CTS\footnotemark[1] \& DToP\footnotemark[1]  & 72.7 & 135 & 22 & 75.0 & 296 & 21  \\
+dCTS \( \tau \)-6899       & 72.8 & 182 & 68 & 72.7 & 247 & 62 \\
+STEP@[6,8]\( \tau \)-6899  & 72.6 & 131 & 21 & 72.0 & 183 & 25 \\
+STEP@[6,8]\( \tau \)-4999  & 69.8 & 95  & 22 & 71.0 & 149 & 25 \\
+STEP@[8]\( \tau \)-6899    & 69.9 & 149 & 28 & 72.0 & 199 & 35 \\
+STEP@[8]\( \tau \)-4999    & 69.9 & 105 & 22 & 71.1 & 163 & 36 \\

\botrule
\end{tabular*}
\footnotetext[1]{Default configuration from the original paper}
\end{table}

The results in Table~\ref{STEP_ViTB_Cityscapes_high} and Table~\ref{STEP_ViTL_Cityscapes_high} highlight how our STEP mechanism and the dCTS patch merger effectively handle varying image resolutions. As the resolution increases to $768\times768$ and $1024\times1024$, SegViT suffers a dramatic increase in computational cost and a substantial drop in inference speed. Our STEP configurations maintain a more stable trade-off. When using ViT-Base as the backbone, we observe that the reduction in FLOPs is most significant compared to the two reference methods, CTS and DToP. However, this does not consistently translate into proportional gains in throughput. In this case, a noticeable drop (up to 4\%) in segmentation quality is observed. This suggests that the confidence threshold used in our early-pruning mechanism may need to be re-evaluated to better balance efficiency and accuracy. The dCTS, particularly its\( \tau \)-6899 variant, emerges as a strong compromise between accuracy and efficiency. It consistently delivers high mIoU across both backbones and resolutions, with much lower GFLOPs and significantly improved throughput. For example, on ViT-Large at $1024\times1024$, dCTS achieves only a marginal drop in mIoU while requiring just 802 GFLOPs, which is 2.6 times less complex than SegViT, and reaches 41 FPS, surpassing SegViT's 12 FPS by more than a factor of three. This demonstrates the capacity of dCTS to maintain segmentation quality while significantly enhancing inference efficiency in high-resolution and real-time applications. By applying the complete STEP framework, computational complexity can be reduced by as much as 4×. Overall, the higher the input image resolution and the larger the backbone, the more our STEP approach exhibits clear advantages in terms of both efficiency and segmentation performance

\begin{table}[htbp]
\caption{Performance evaluation of our STEP mechanism integrated with ViT-Large.}\label{STEP_ViTL_Cityscapes_high}
\begin{tabular*}{\textwidth}{@{\extracolsep\fill}lcccccc}
\toprule%
& \multicolumn{3}{@{}c@{}}{Cityscapes ($768\times768$)} & \multicolumn{3}{@{}c@{}}{Cityscapes ($1024\times1024$)} \\\cmidrule{2-4}\cmidrule{5-7}%
Method & mIoU$\nearrow$ & GFLOPs$\searrow$ & FPS$\nearrow$ & mIoU$\nearrow$ & GFLOPs$\searrow$ & FPS$\nearrow$ \\
\midrule
SegViT                      & 74.4 & 970 & 37 & 75.7 & 2086 & 12\\
CTS\footnotemark[1]         & 74.5 & 622 & 40.5 & 75.7 & 1283 & 20.5\\
DToP\footnotemark[1]        & 73.7 & 589 & 13 & 75.4 & 1176 & 7.5\\
\hline
\hline
+dCTS\( \tau \)-6899        & 74.4 & 598 & 47  & 74.9 & 802 & 41\\
+STEP@[8,16]\( \tau \)-6899 & 73.6 & 424 & 13  & 73.8 & 514 & 13.5\\
+STEP@[18]\( \tau \)-6899   & 74.3 & 490 & 23.5 & 74.5 & 655 & 20.5\\

\botrule
\end{tabular*}
\footnotetext[1]{Default configuration from the original paper}
\end{table}

\begin{figure}[htbp]
\centering
\includegraphics[width=\textwidth]{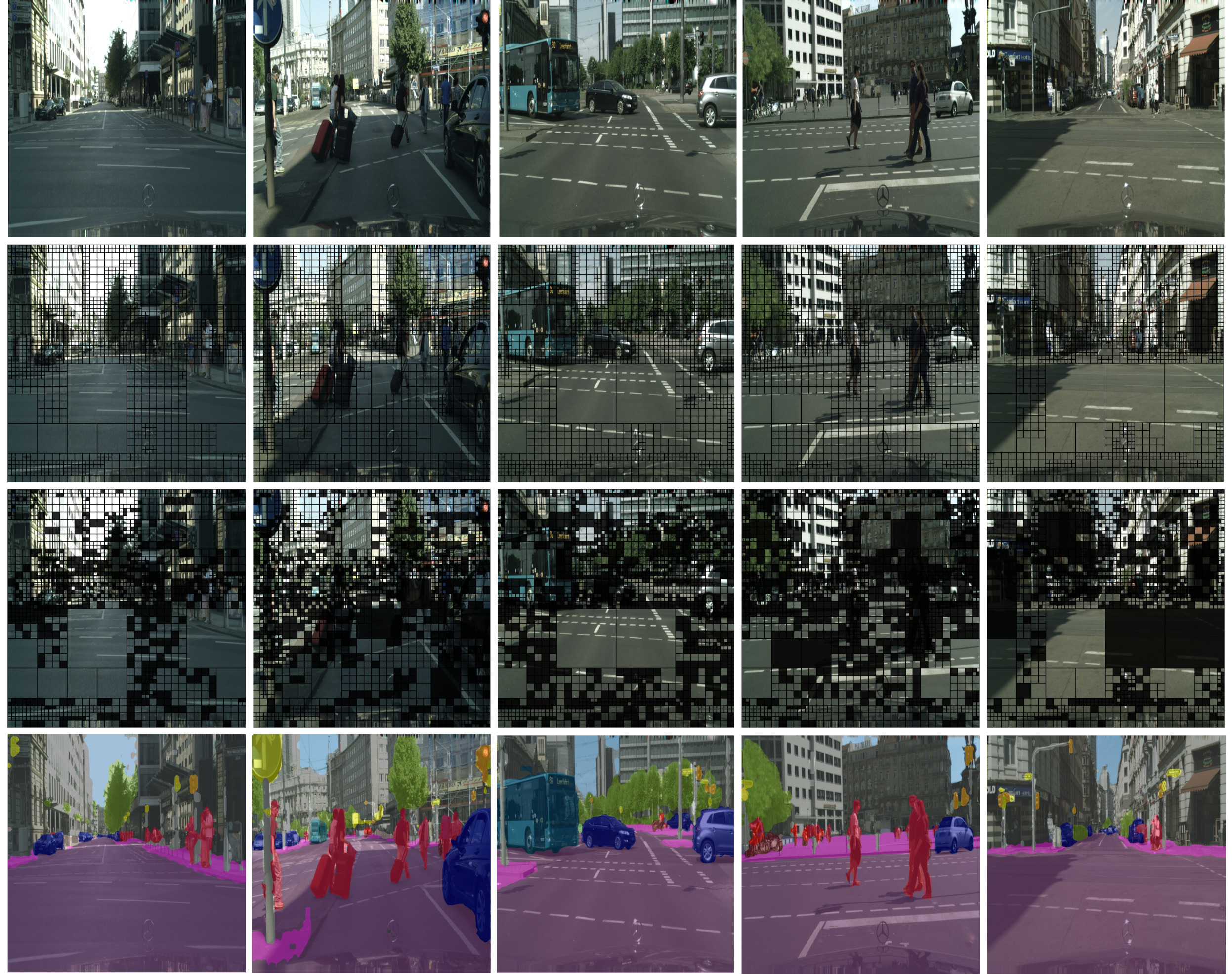}
\caption{Distribution of pruned supertokens across the different stages using STEP@[8] on ViT-Base. From top to bottom: input image from the Cityscapes dataset at $1024\times1024$ resolution, generated superpaches via dCTS, pruned tokens marked in black, and final segmentation results.}
\label{Cityscapes_VitB_1024}
\end{figure}

\begin{figure}[htbp]
\centering
\includegraphics[width=\textwidth]{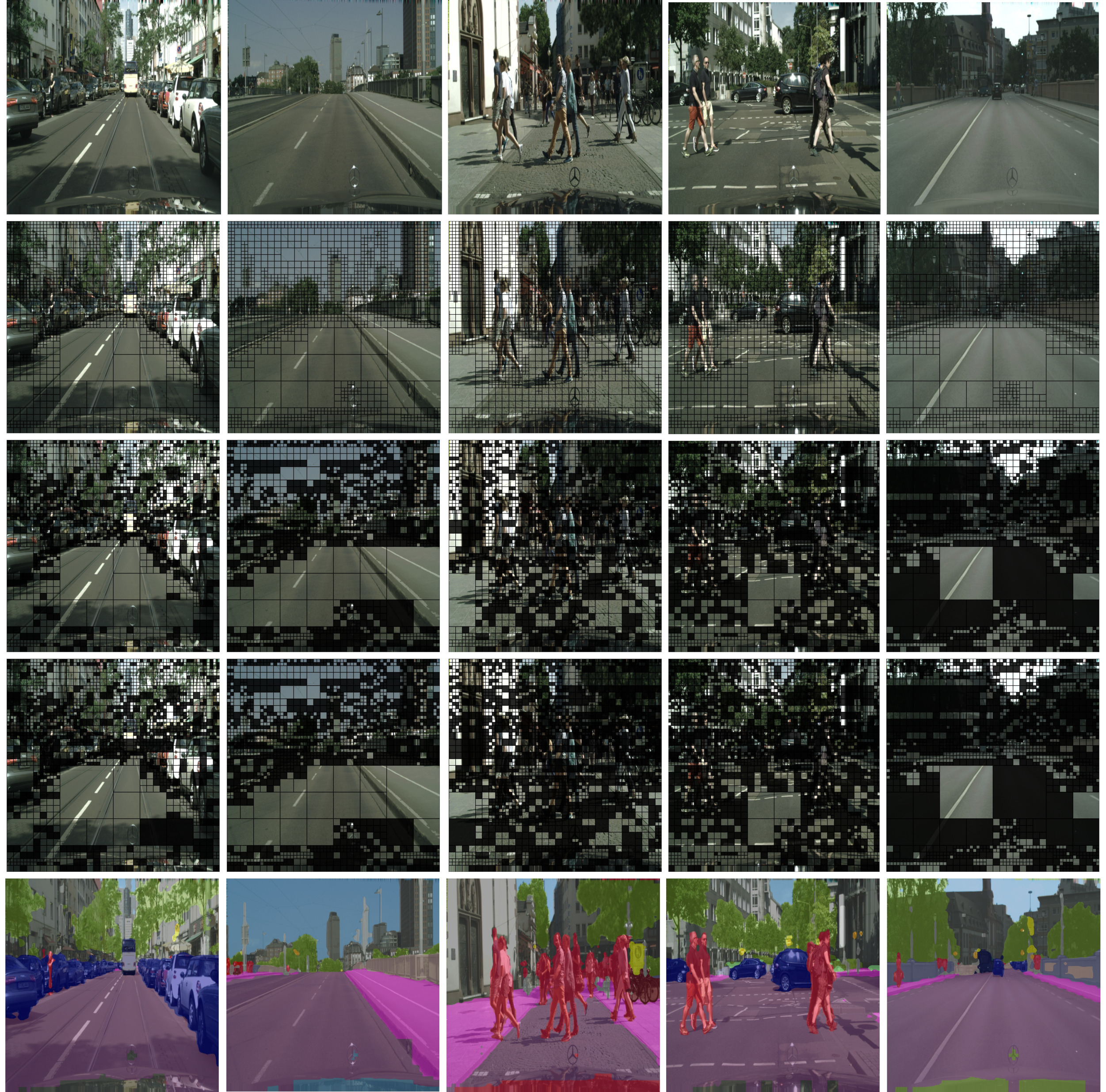}
\caption{Distribution of pruned supertokens across the different stages using STEP@[8,16] on ViT-Large. From top to bottom: input image from the Cityscapes dataset at $1024\times1024$ resolution, generated superpaches via dCTS, pruned tokens marked in black for auxiliary heads 1 and 2, and final segmentation results.}
\label{Cityscapes_VitL_1024}
\end{figure}

We observe that strategically placing the pruning heads allows for greater reduction in GFLOPs. This is likely due to the fact that, regardless of the configuration, an average 48\% of tokens can be halted early in the network for ViT-Large under high-resolution images. Adding STEP to ViT-B when processing standard-resolution images results in an average pruning of 39\% of tokens (Table~\ref{DTOP_stats_merged}). This is a consistent trend, which we also observed in the ablation study (\ref{DToP_config}) conducted on COCOStuff10k at the same resolution. Figure~\ref{Cityscapes_VitL_1024} and Figure~\ref{Cityscapes_VitB_1024} illustrates how tokens are halted across images by each auxiliary head, revealing that many tokens are pruned early in simple scenarios, while they are retained until the final prediction phase in more complex scenes.

\begin{table}[htbp]
\caption{Token pruning dynamics per auxiliary head in STEP on Cityscapes for various input resolutions.}
\label{DTOP_stats_merged}
\centering
\begin{tabular}{l cc cc cc}
\toprule
& \multicolumn{3}{c}{Input $512\times512$ ViT Base} & \multicolumn{3}{c}{Input $1024\times1024$ ViT Large} \\
\cmidrule(lr){2-4} \cmidrule(lr){5-7}
& \multicolumn{2}{c}{STEP@[6,8]} & STEP@[8] & \multicolumn{2}{c}{STEP@[8,16]} & STEP@[18] \\
\cmidrule(lr){2-3} \cmidrule(lr){4-4} \cmidrule(lr){5-6} \cmidrule(lr){7-7}
After & Aux1 & Aux2 &  Aux1 &  Aux1 &  Aux2 &  Aux1 \\
\midrule
Mean    & 245 & 269 & 273 & 717  & 885  & 833 \\
Maximum & 400 & 422 & 446 & 1143 & 1340 & 1238 \\
Minimum & 107 & 127 & 101 & 300  & 362  & 247 \\
\bottomrule
\end{tabular}
\end{table}

\begin{figure}[htbp]
    \centering
    \begin{subfigure}[b]{1.0\textwidth}
        \centering
        \includegraphics[width=\textwidth]{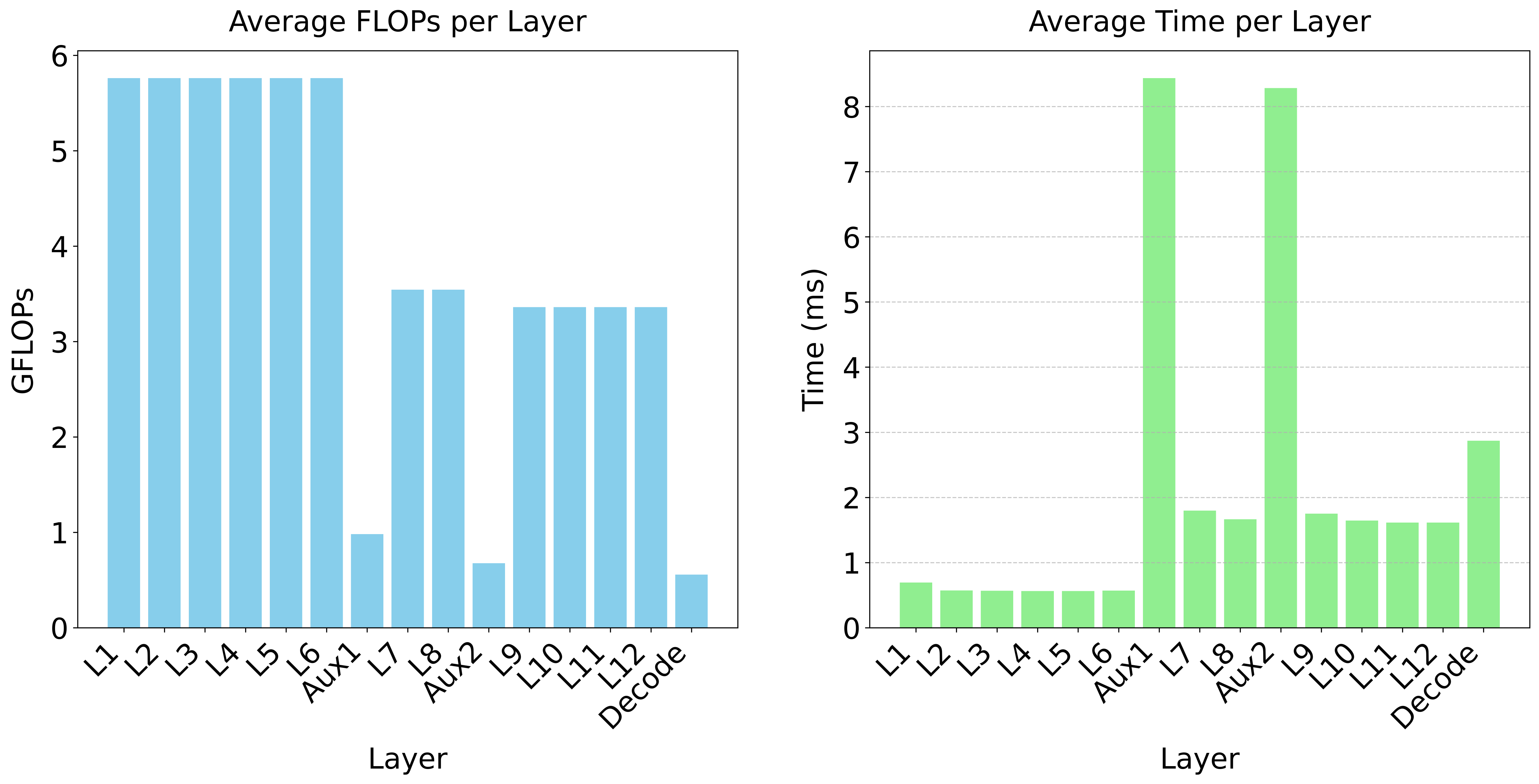}
        \caption{STEP@[6,8]}
    \end{subfigure}
    \vskip\baselineskip
    \begin{subfigure}[b]{\textwidth}
        \centering
        \includegraphics[width=\textwidth]{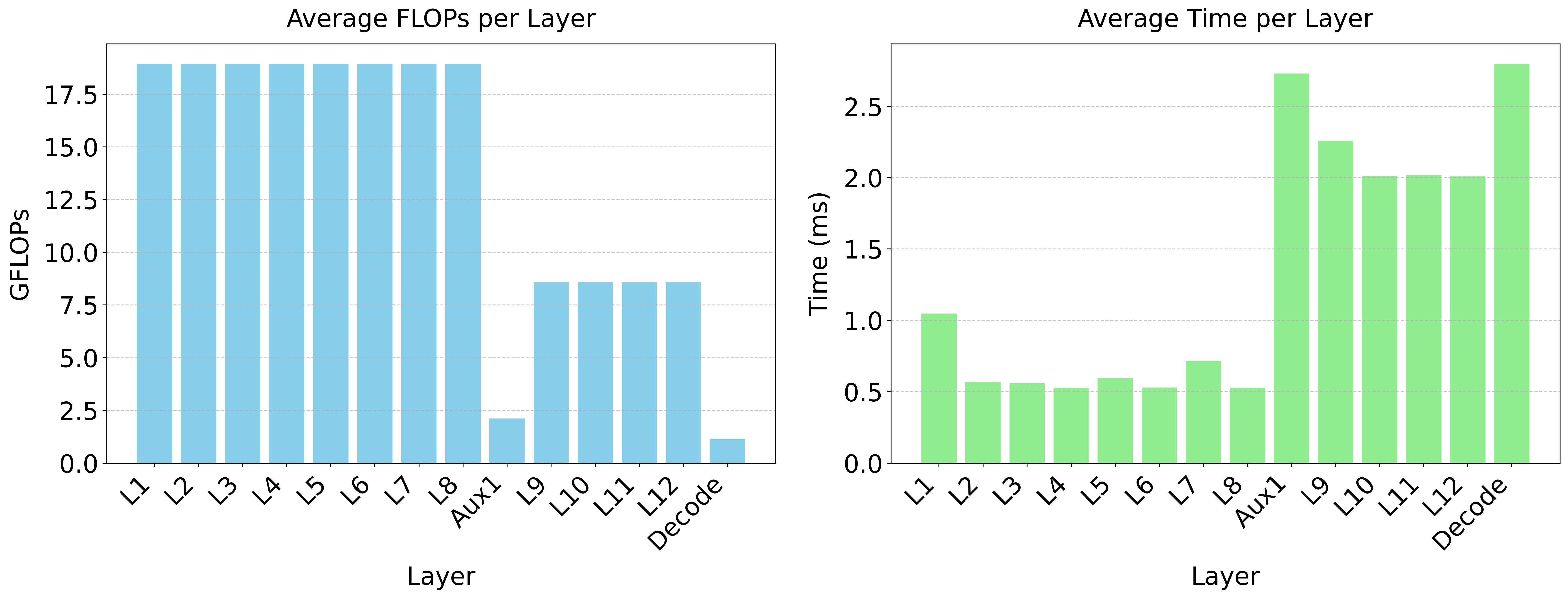}
        \caption{STEP@[8]}
    \end{subfigure}
    
    \caption{Per-layer computational complexity (GFLOPs) and throughput (FPS) analysis across encoder and auxiliary pruning heads using ViT-Base as backbone on the Cityscapes dataset at 1024×1024 resolution.}
    \label{fig:ViTB_FLOPS_FPS_City1024x1024}
\end{figure}

\begin{figure}[htbp]
    \centering
    \begin{subfigure}[b]{1.0\textwidth}
        \centering
        \includegraphics[width=\textwidth]{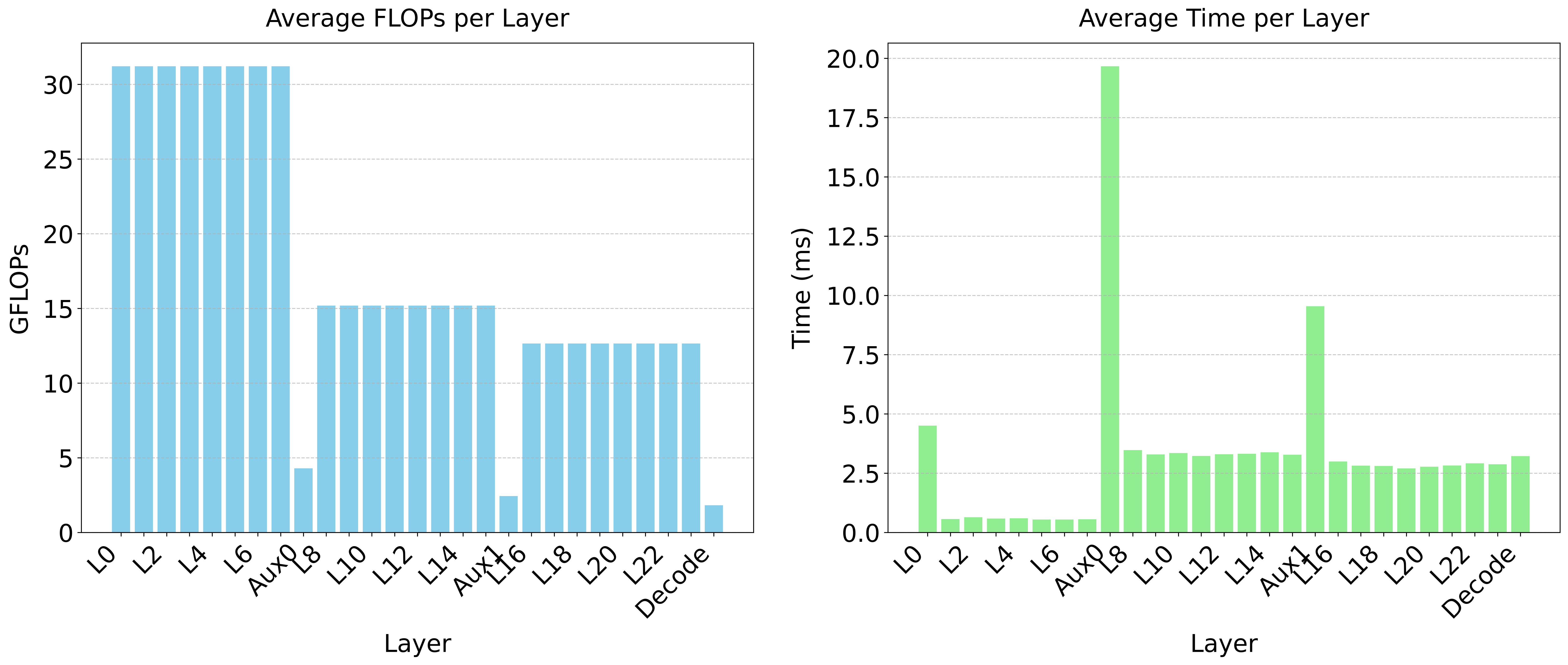}
        \caption{STEP@[8,16]}
    \end{subfigure}
    \vskip\baselineskip
    \begin{subfigure}[b]{\textwidth}
        \centering
        \includegraphics[width=\textwidth]{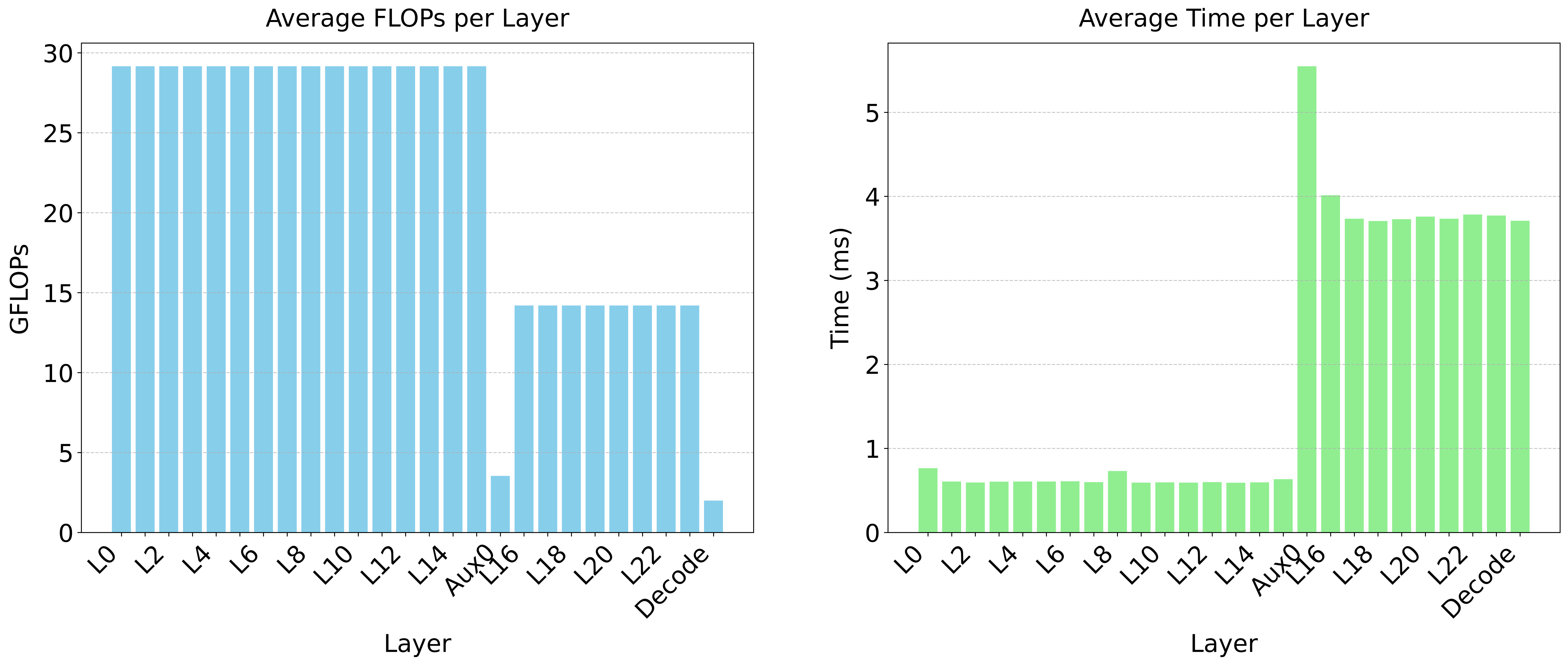}
        \caption{STEP@[18]}
    \end{subfigure}
    
    \caption{Per-layer computational complexity (GFLOPs) and throughput (FPS) analysis across encoder and auxiliary pruning heads using ViT-Large as backbone on the Cityscapes dataset at 1024×1024 resolution.}
    \label{fig:ViTL_FLOPS_FPS_City1024x1024}
\end{figure}

To better understand why total throughput does not scale linearly with FLOPs savings, we present Figure~\ref{fig:ViTB_FLOPS_FPS_City1024x1024} and Figure~\ref{fig:ViTL_FLOPS_FPS_City1024x1024}, which show the per-layer inference time and computational cost with STEP applied to both ViT-Base and ViT-Large under high-resolution. Although fewer tokens are processed in the later layers, identifying and discarding high-confidence tokens introduces bottlenecks. Despite the low computational cost of the auxiliary pruning head based on the ATM module, the masking operations likely cause the observed slowdown due to their irregular control flow, dynamic memory access patterns, and tensor shape variability. Further analyse of computation flow is necessary to verify if all operations are realized on GPU and there is unnecessary memory copy between GPU and CPU which can effectively slow the whole process. Additional overheads, such as memory allocation and synchronization costs, may further diminish the expected performance gains. These findings suggest that optimizing only the number of tokens is insufficient, one must also consider the computational efficiency of the pruning mechanism itself.

\section{Conclusion}\label{Conclusion}
We introduced a novel token reduction method, SuperToken and Early-Pruning (STEP), designed to improve token efficiency in ViTs for semantic segmentation. STEP combines adaptive patch merging with an early-pruning mechanism. At the core of this method lies an enhanced patch-level merging technique, referred to as dCTS, which employs a flexible strategy to form square-shaped superpatches of varying sizes, allowing the model to better capture the spatial complexity of image content. Additionally, we investigated the benefits of early-pruning tokens via DToP within the network. Our experiments were conducted under varying image resolution settings, encompassing both low and high-resolution inputs. To the best of our knowledge, this is the first work to systematically assess the effect of token pruning across different resolution levels. STEP demonstrated strong scalability on both ViT-Base and ViT-Large with high-resolution images, offering substantial computational savings while preserving most of the segmentation accuracy, making it a compelling choice for efficient dense prediction in high-resolution scenarios. However, this efficiency gain does not always resulted in a proportional throughput improvement. The dCTS alone  showed particularly strong robustness to increasing input resolutions. Across all configurations, a small but consistent drop in mIoU was observed compared to the baseline, suggesting that some relevant tokens may have been prematurely discarded or merged. This highlights the importance of carefully tuning the fusion thresholds within STEP especially when operating under high-resolution regimes. The early pruning mechanism using ATM-based auxiliary heads allows up to 48\% of tokens to be halted. While this further reduces computational complexity, it significantly slows down inference, regardless of image resolution. This is likely due to hardware inefficiencies introduced by the masking mechanism. Further work could focus on analyzing current solution and improving algorithm design to better leverage GPU parallelism and avoid GPU-to-CPU data transfers.

\backmatter

\bibliographystyle{sn-mathphys-num}


\end{document}